\documentclass[twoside,11pt]{article}

\usepackage{jmlr2e}
\usepackage{amsmath,amssymb,graphicx,bbm,enumerate}
\usepackage{color}
\usepackage[dvipsnames]{xcolor}
\usepackage{mathrsfs}
\usepackage{url}
\usepackage[normalem]{ulem}

\def\cN{{\mathcal N}} 

\def\hat#1{\widehat{#1}} 
\def\wt#1{\widetilde{#1}} 

\def\bd#1{\boldsymbol{#1}}

\def\argmax{\operatornamewithlimits{arg\,max}}

\def\bi{\begin{itemize}}
\def\ei{\end{itemize}}
\def\be{\begin{enumerate}}
\def\ee{\end{enumerate}}

\newcommand{\eq}[1]{\begin{align*}#1\end{align*}}
\newcommand{\eqn}[1]{\begin{align}#1\end{align}}
\newcommand{\indep}{\rotatebox[origin=c]{90}{$\models$}}
\newtheorem{prop}{Proposition}

\graphicspath{{./figures/}}

\jmlrheading{1}{2000}{1-48}{4/00}{10/00}{meila00a}{Shaobo Han and David B. Dunson}

\ShortHeadings{Supervised Multiscale Dimension Reduction}{Han and Dunson}
\firstpageno{1}

\begin{document}

\title{Supervised Multiscale Dimension Reduction for Spatial Interaction Networks}

\author{\name Shaobo Han \email shaobohan@gmail.com \\
       \name David B.\ Dunson \email dunson@duke.edu \\
       \addr Department of Statistical Science\\
       Duke University\\
       Durham, NC 27708-0251, USA}

\editor{}

\maketitle
\begin{abstract}
{\color{black}
We introduce a multiscale supervised dimension reduction method for SPatial Interaction Network (SPIN) data, which consist of a collection of spatially coordinated interactions. This type of predictor arises when the sampling unit of data is composed of a collection of primitive variables,  each of them being  essentially unique, so that it becomes necessary to group the variables in order to simplify the representation and enhance interpretability. In this paper, we introduce an empirical Bayes approach called \emph{spinlets}, which first constructs a partitioning tree to guide the reduction over multiple spatial granularities, and then refines the representation of predictors according to the relevance to the response. We consider an inverse Poisson regression model and propose a new multiscale generalized double Pareto prior, which is induced via a tree-structured parameter expansion scheme. Our approach is motivated by an application in soccer analytics,  in which we obtain compact vectorial representations and readily interpretable visualizations of the complex network objects, supervised by the response of interest.  
}

\end{abstract}


\section{Introduction}

In modern applications, we frequently encounter complex object-type data, such as functions \citep{ramsay2006functional},  trees \citep{wang2007object},    shapes \citep{srivastava2011shape}, and networks \citep{durante2017nonparametric}.  
In many instances, such data are collected repeatedly under different conditions, with an additional response variable of interest available for each replicate. This has motivated an increasingly rich literature on generalizing regression on vector predictors to settings involving more elaborate object-type predictors with special characteristics, such as functions \citep{james2002generalized}, manifolds \citep{nilsson2007regression},  tensors \citep{zhou2013tensor}, and undirected networks \citep{guha2018bayesian}. 

\begin{figure}[hbpt] 
\vskip 0.00in
\begin{center}
\footnotesize{
$\begin{array}{cc}
\hspace{-0.3cm}\includegraphics[height=0.34\textwidth, width=0.38\textwidth]{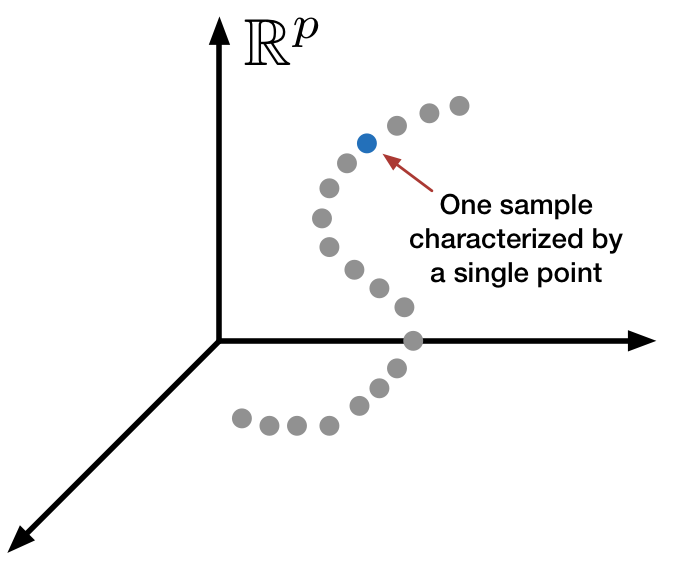} & \hspace{-0.0cm}\includegraphics[height=0.34\textwidth, width=0.38\textwidth]{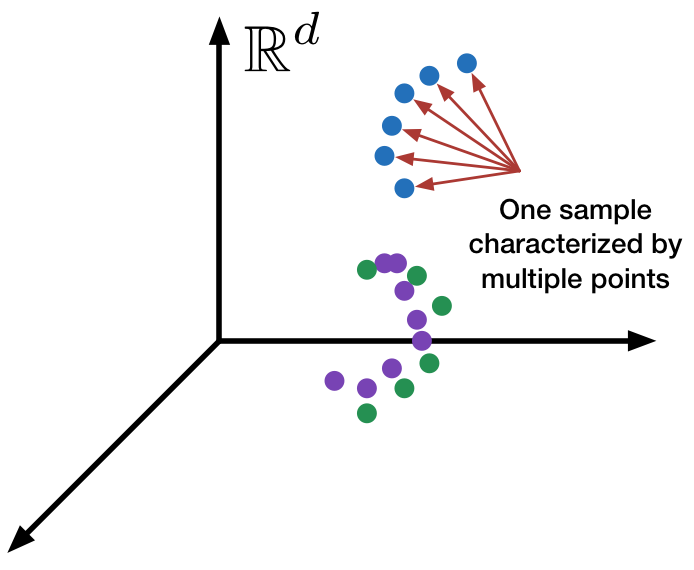}\\
\hspace{-0.3cm} \textrm{(a) } & \hspace{-0.0cm} \textrm{(b)}  \\
\end{array}$}
\end{center}
 \vskip -0.10in
\caption{{\color{black}Comparison of data sampling unit in $(a)$ the conventional regression setting and $(b)$ the composite object regression setting. The composite objects in green color and purple color in $(b)$ are apparently similar to each other, but the total number of component primitive objects are different.}
}
\label{fig0: dimension_reduction}
\end{figure}

Complex objects are often built from simpler parts. {\color{black}In this article, we consider supervised dimension reduction in which the sampling unit of data is in the form of \emph{composite objects} (CO), composed of a collection of \emph{primitive objects} (POs). Many data types can be seen as instances of the CO family, such as networks with spatially coordinated edges, pathology images of cells with a variety of shaped nuclei, or microbiome samples with species related on a phylogenetic tree. As shown in Figure \ref{fig0: dimension_reduction},  in the conventional regression formulation with a vector predictor, each replicate corresponds to a single point in  $\mathbb{R}^{p}$. In regression with a composite object-valued predictor, the data could lie in a relatively lower dimensional $\mathbb{R}^{d}$ space, but each replicate corresponds to a collection of points. The component POs in CO-type data can be  enormous and mostly distinctive from one another across replicates,  presenting new challenges for data exploration, analysis, and visualization. 

}

\begin{figure}[hbpt] 
\vskip 0.00in
\begin{center}
\footnotesize{
$\begin{array}{cc}
\hspace{-0.3cm}\includegraphics[height=0.34\textwidth, width=0.49\textwidth]{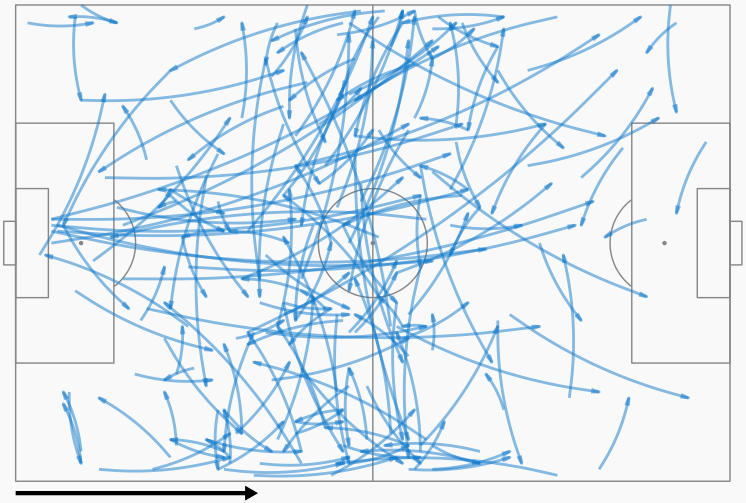} & \hspace{-0.3cm}\includegraphics[height=0.34\textwidth, width=0.49\textwidth]{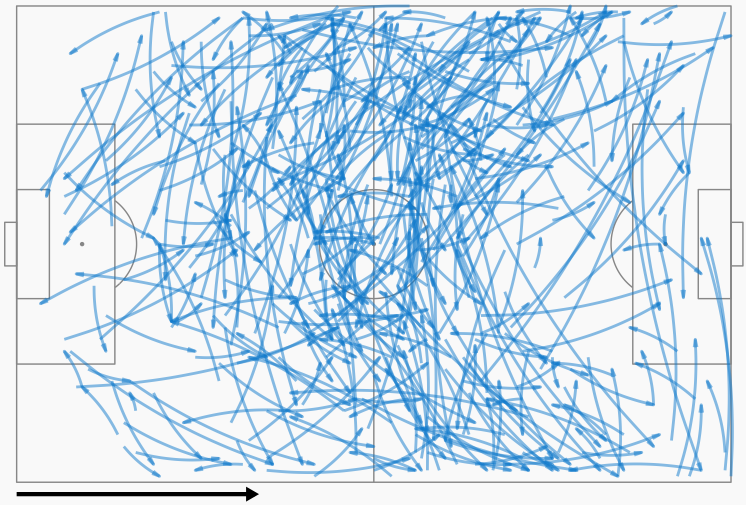}\\
\hspace{-0.1cm} \textrm{(a) $202$ completed passes by France} & \hspace{-0.1cm} \textrm{(b) $448$ completed passes by Croatia}  \\
\end{array}$}
\end{center}
 \vskip -0.10in
\caption{Spatial interaction networks in the 2018 FIFA World Cup Final
 (France 4-2 Croatia). The arrowed segments denote the pass from the location of passer to the location of receiver. Team's  direction of attack: from left to right. 
}
\label{fig1: france-croatia}
\end{figure}

{\color{black}We are motivated by the 2018 FIFA World Cup data collected by StatsBomb (\href{https://statsbomb.com/}{\url{https://statsbomb.com/}}), which contain information on the association between coordinated interactions between players and team performance. For each replicate $i \in \{1,\hdots, n\}$, we observe SPatial Interaction Networks (SPIN) data $\mathcal{E}_{i}:=\{e_{k}: k=1,\hdots, q_{i}\}$, which contains a collection of $q_{i}$ completed passes.}  As illustrated in Figure \ref{fig1: france-croatia}, $\mathcal{E}_{i}$ is in the CO form, with every constituent pass $e_{k}$ viewed as a PO logged with the spatial locations of the passer and receiver. {\color{black}The whole dataset contains  $50,159$ completed passes with $49,988$ unique locations of origin-destination in the $64$ matches.} The associated response $y_{i}$ may refer to a team performance metric such as goals scored or conceded,  the number of shots on target, or other situational game factors. 

We consider regression modeling with $n$ observations of a scalar response ${y}_{i}$  and {\color{black}SPIN as} a CO-valued predictor $\mathcal{E}_{i}$, $i = 1, \hdots, n$. A primary challenge is to represent the CO data in a malleable form that facilitates multivariate analysis.  One common practice is to divide $\mathcal{P}:=\bigcup_{i=1}^{n} \mathcal{E}_{i}$, the complete set of POs, into $p$ non-overlapping subsets $\mathcal{P}={\bigcup}_{j=1}^{m}\Pi_{j}$ (with $\Pi_{j}\bigcap\Pi_{j'} = \emptyset$ for $j, j'\in \{1, \hdots, p\}$ and $j\neq j'$) through a predefined partitioning scheme $\bd{\Pi}=[\Pi_{1}, \hdots, \Pi_{p}]$. Under the partition $\bd{\Pi}$, the $\mathcal{E}_{i}$ can be represented as a $p$-dimensional count vector $\bd{x}_{i} \in\mathbb{Z}^{p}$ {\color{black}through a \emph{bag-of-words} representation \citep{blei2003latent, taddy2013multinomial}}, where $x_{i, j} = N(\mathcal{E}_{i} \bigcap {\Pi}_{j})$ counts the occurrences of  POs appearing in $\mathcal{E}_{i}$ and belonging to subset $\Pi_{j}$.  

\begin{figure}[hbpt] 
\vskip 0.00in
\begin{center}
\footnotesize{
$\begin{array}{cc}
\hspace{-0.3cm}\includegraphics[height=0.34\textwidth, width=0.49\textwidth]{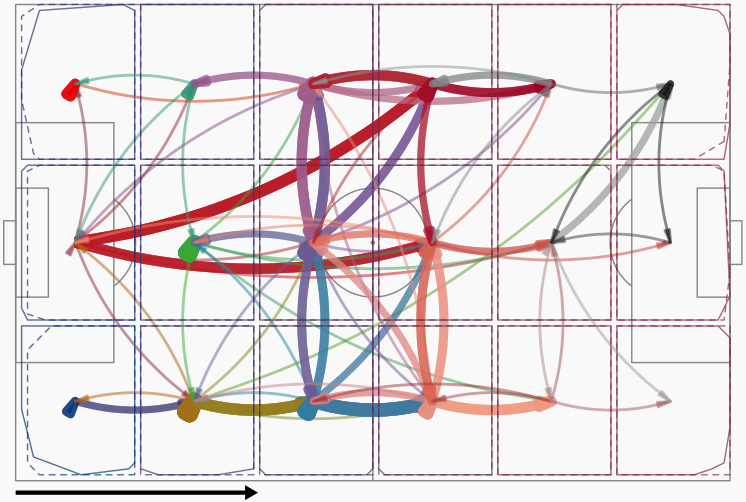} & \hspace{-0.3cm}\includegraphics[height=0.34\textwidth, width=0.49\textwidth]{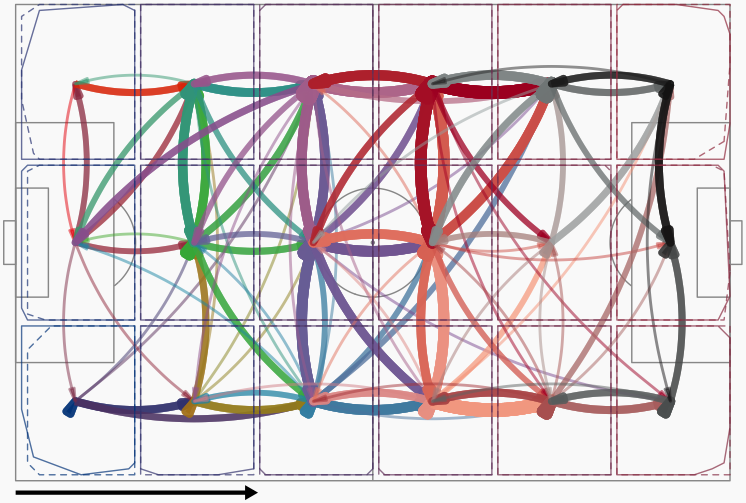}\\
\hspace{-0.1cm} \textrm{(a) Zone passing network of France} & \hspace{-0.1cm} \textrm{(b) Zone passing network of Croatia}  \\
\end{array}$}
\end{center}
 \vskip -0.10in
\caption{Tile-based network representation induced by a predefined $3\times6$ uniform parcellation of the pitch. The colored arrows represent the grouped POs with width proportional to the count of occurrences in the group. 
}
\label{fig2: viz}
\end{figure}

For example, \cite{miller2014factorized} discretizes the basketball court uniformly into tiles and counts the number of shots located in each tile. \cite{durante2017nonparametric} parcellates the brain into regions and investigates the network connectivity between pairs of regions. {\color{black}In Figure \ref{fig2: viz}, we plot the resulting tile-based network representation based on a \emph{nodal} partitioning scheme. Unfortunately, such a non-data adaptive scheme throws away many relevant details and produces an overly sparse representation of the data.}

{\color{black}Alternatively, one can directly divide POs into disjoint groups according to their similarity.} This partition scheme inherently assumes the equivalence of the POs falling within the same group, and focuses on the variabilities in abundance across groups. The choice of partitioning scheme and its scale will have a critical influence on inference.  Ideally, the induced vector representation should promote the interpretability of the CO-type predictor and preserve the relevance to the response. However, such approaches are underdeveloped in the current literature.

{\color{black} In this article, we propose \emph{spinlets}---a supervised multiscale dimension reduction method for spatial interaction networks and CO-type data more broadly. Spinlets takes into account both the predictor structure and the label supervision: it uses the similarity among POs for deriving a \emph{bag-of-POs} representation accompanied by a partitioning tree, and then refine the tree structure according to the relevance to the response. Spinlets serves as a tool for both supervised representation learning and visualization of complex SPIN data. We first review the relevant literature and then describe our spinlets method.}  
\subsection{Relevant Literature}
There is a rich literature on supervised dimension reduction, covering LASSO \citep{tibshirani1996regression}, supervised PCA \citep{bair2006prediction, barshan2011supervised}, and sufficient dimension reduction (SDR), see \cite{cook2007fisher}, \cite{adragni2009sufficient} and the references cited therein. The reduction of complexity is typically achieved through variable selection or combination.  Such approaches can accommodate vector predictors and perform well in high-dimensional settings; however, our application involves predictors with complex structures. SDR methods have been generalized to handle functional predictors \citep{ferre2003functional, ferre2005smoothed}, matrix- or array-valued predictors \citep{li2010dimension}, and irregularly measured longitudinal predictors \citep{jiang2014inverse}. In this article, we center our focus on spatial networks as an instance of a composite data object. 

There is a separate literature on multiscale geometric data representation, including diffusion maps \citep{lafon2006diffusion} and GMRA \citep{allard2012multi, petralia2013multiscale}. These approaches seek a reductive representation that reflects the intrinsic geometry of the high-dimensional data by partitioning the similarity graph of $n$ data observations. In contrast, our spinlets method partitions the similarity graph of  $q= \sum_{i=1}^{n}q_{i}$ variables, with a different goal of identifying predictive groups of variables. Spinlets is similar in spirit with the treelets method \citep{lee2008}, which organizes variables on a hierarchical cluster tree with multiscale bases; however, treelets is an unsupervised approach utilizing the sample correlation to construct the tree with a single cutoff height. Our spinlets approach departs from treelets by incorporating external proximity to construct the tree, and determining non-uniform heights \citep{meinshausen2008discussion} with reference to the response. 

In regression with CO-type predictors, the total number of unique POs is massive, while only a limited number of them are sparsely observed within each replicate. It is advantageous to form groups of POs that are spatially contiguous, such that meaningful analysis can be conducted at a lower level of resolution. In many other applications, predictors are highly correlated, or collectively associated with the response, or domain knowledge exists suggesting the functional similarity among a group of variables. This has motivated a line of research on supervised clustering of predictors in forward regression settings. Examples include \cite{hastie2001supervised, jornsten2003simultaneous} and \cite{dettling2004finding}. The averaging operator on the predictors often leads to lower variance \citep{park2006averaged}.   

Regularization methods such as elastic net \citep{zou2005regularization} or OSCAR \citep{bondell2008simultaneous} can mitigate the multicollinearity issue and encourage grouping effects. Along this thread, \cite{wang2017constructing} recently proposed two tree-guide fused lasso (TFL) penalties, which effectively encode the topology of a phylogenetic tree in determining the taxonomic levels of microbes associated with a given phenotype. However, this approach does not model the variability in the predictors,  while we model the conditional distributions of the predictors given the response through inverse regression, with possibilities of alleviating the effects of collinearity \citep{cook2009dimension}. Moreover, Lasso-based penalties tend to over-shrink signals not close to zero \citep{armagan2011generalized}. We introduce a new multiscale prior that induces a locally adaptive shrinkage rule on each scale. {\color{black}This prior executes two special operators on the tree within our proposed spinlets method.}

In Section \ref{sec3}, we introduce {\color{black}the Poisson inverse regression model and the reductive operators.} Section \ref{sec4} presents a tree-structured PX scheme and our new multiscale shrinkage prior. A variational expectation-maximization (EM) algorithm for estimation is outlined in Section \ref{sec5}.  In Section \ref{sec6},  we evaluate the performance of our approach with simulated data and demonstrate the practical utility through applications to soccer analytics.  The implementation of spinlets will be made available on Github.

\section{Tree-Guided Supervised Dimension Reduction }\label{sec3}

{\color{black}As a preprocessing step, we build a partition tree via recursively applying the METIS partitioning algorithm \citep{karypis1998fast} on the similarity graph $\mathcal{G}$ of POs, detailed in Appendix \ref{appendix1}. The multiscale proximity information of POs is encoded in a binary tree $\mathcal{T}_{h}$ of height $h$, which partitions $\mathcal{P}$ into $m = 2^{h}$ {groups} on the finest scale and yields a primary vectorial representation $\bd{X} \in\mathbb{Z}^{m}$ of COs, aligned across replicates.} 

According to \cite{cook2007fisher}, a sufficient reduction of the random vector $\bd{X}$, denoted as $R(\bd{X})$, satisfies one of the  three equivalent statements: $(i)$  $\bd{X}|(Y, R(\bd{X}))\sim \bd{X}|R(\bd{X})$; $(ii)$ $Y|\bd{X}\sim Y|R(\bd{X})$; $(iii)$ $\bd{X}\indep Y|R(\bd{X})$, where $\sim$ indicates equivalence in distribution and $\indep$ denotes independence. Our main goal is to determine a reductive rule $R(\bd{X})$ by pruning the tree $\mathcal{T}_{h}$, such that the resulting representation is interpretable and retains the relevance to the response variable $Y$.

\subsection{Poisson Inverse Regression Model}\label{subsec:PIR}

 For replicate $i$ with response variable $Y_{i}$, we attach a random variable $X_{i,j}\in\mathbb{Z}$ to each leaf node $j$, counting the number of POs appearing in CO $\mathcal{E}_{i}$ that fall in the {\color{black}$j$-th} leaf group. In order to explicitly model the variabilities in occurrences, we adopt an \emph{inverse regression} formulation. This approach is motivated by poor performance we observed in implementing usual regression due to extreme multicollinearity issues. Sufficiency is guaranteed within our proposed Poisson inverse regression (PIR) model  for  $X_{i,j}$ conditionally on $Y_{i}$, $i=1,\ldots, n$, $j = 1,\ldots, m$, 
\eqn{
(X_{i,j}|Y_{i}=y_{i})\sim \textrm{Poisson}(\lambda_{i, j}), \quad \eta_{i, j}(y_{i})=\ln(\lambda_{i, j}) = \alpha_{j} + \mu_{i}+y_{i}{\beta}_{j} ,\label{eq:model}
}
where $\alpha_{j}$ is the intercept for predictor $j$,  $\mu_{i}$ is the baseline effect for replicate $i$, and ${\beta}_{j}$ is the regression coefficient for predictor $j$. The linear sufficient reduction $R_{\bd{\beta}}(\bd{X})$ parameterized by $\bd{\beta}$ is derived as follows (the replicate index $i$ is omitted for clarity).

\begin{prop}
Letting $R_{\bd{\beta}}(\bd{X})=\bd{\beta}^{T}\bd{X}$,  under the inverse Poisson regression model \eqref{eq:model}, the distribution of $Y|\bd{X}$ is the same as the distribution of $Y|R_{\bd{\beta}}(\bd{X})$ for all values of $\bd{X}$. 
 \end{prop}
 \cite{cook2007fisher} proves the sufficiency of $R_{\bd{\beta}}(\bd{X})=\bd{\beta}^{T}\bd{X}$ for one-parameter exponential family models. Within this family, the PIR model \eqref{eq:model} can be written in the following form,
\eq{
f_{j}(x_{j}|Y=y) & = a_{j}(\eta_{j}(y))b_{j}(x_{j})\exp{(x_{j}\eta_{j}(y))}, \cr a_{j}(\eta_{j}(y)) & = \exp{[{-\exp{({\eta_{j}(y)}})}]}, \quad b_{j}(x_{j})=1/x_{j}! 
}
Accordingly, the joint probability mass function  $f(\bd{x}|y)$ of $\bd{X}|(Y=y)$ can be written as 
\eq{f(\bd{x}|y) = g(\bd{\beta}^{T}\bd{x}, y) h(\bd{x}),}
 where $g(\bd{\beta}^{T}\bd{x}, y) = \exp{(y\bd{\beta}^{T}\bd{x})}\prod_{j=1}^{m} a_{j}(\eta_{j}(y))$ and $h(\bd{x}) = \prod_{j=1}^{m} [b_{j}(x_{j})\exp{({x}_{j}{\alpha}_{j})}]$. Thus, the sufficiency of {\color{black}linear} reduction holds according to the Fisher-Neyman factorization theorem for sufficient statistics \citep[Theorem 1.5.1, p. 43]{bickel2015mathematical}. The PIR model has a close connection with the multinomial inverse regression (MNIR) model \citep{taddy2013multinomial}, though, the vector Poisson likelihood departs from multinomial likelihood by accounting for the variability of total number of POs in each replicate.   
 
 \subsection{Reductive Operators: Deletion and Fusion}\label{sec: operators}

One reductive operator enabled by the $\bd{\beta}$ parameterization from the PIR model \eqref{eq:model} is the \emph{deletion} of irrelevant leaf groups. One can see that  $\beta_{j}=0$ implies that $f(x_{j}|Y=y) \equiv f(x_{j})$, that is, the number of POs in the {\color{black}$j$-th} leaf group is independent of the value of  $Y$. Another reductive operator on the tree is \emph{fusion}. We observe that if $\beta_{j}=\beta_{j'}\neq 0$, $\forall j, j'\in \mathcal{D}$, then $R_{\bd{\beta}}(\bd{x})$ is a function depending on the predictors only through $\sum_{j\in \mathcal{D}}{X}_{j}$, which is the total number of POs falling into the set $\mathcal{D}$. Therefore, the practitioner can construct a lower resolution vectorial representation by merging the involved leaf sets into one set $\mathcal{D}$ without loss of relevant information. The relevant signals are then captured on coarser scales.  To ensure spatial contiguity, we require all leaves contained in $\mathcal{D}$ to share at least one common ancestor node.  

The grouping of highly correlated predictors in high-dimensional regression can be incorporated via \emph{fusion penalties} \citep{tibshirani2005sparsity,  bondell2008simultaneous}, which encourage sparsity in the differences of coefficients. There exist several generalized fusion schemes that can take into account graph structure \citep{she2010sparse, tibshirani2011solution}. However, these methods do not support multiscale nested grouping of predictors in accordance with the tree structure. For example, applying the pairwise fused lasso \citep{she2010sparse} to all pairs of variables tends to incorrectly encourage merging all the variables together with equal strengths. The TFL penalties \citep{wang2017constructing} require careful tuning of the regularization parameters across multiple scales. {\color{black}There exists no shrinkage prior that supports the prior belief of sparsity and piece-wise smoothness across scales and locations along the tree.}

\section{Multiscale Shrinkage with Parameter Expansion}\label{sec4}

Parameter expansion (PX) \citep{liu1998parameter} has been found useful not only for accelerating computations \citep{liu1999parameter}, but also for inducing new families of prior distributions \citep{gelman2004parameterization}. In this section, we propose a new tree-structured PX scheme, which induces a multiscale shrinkage prior {\color{black}on $\bd{\beta}$}.

\subsection{Tree Structured Parameter Expansion}\label{sec: tree}
\begin{figure}[hbpt] 
\vskip -0.00in
\begin{center}
{
$\begin{array}{c}
\hspace{-0.3cm}\includegraphics[height=0.33\textwidth, width=0.95\textwidth]{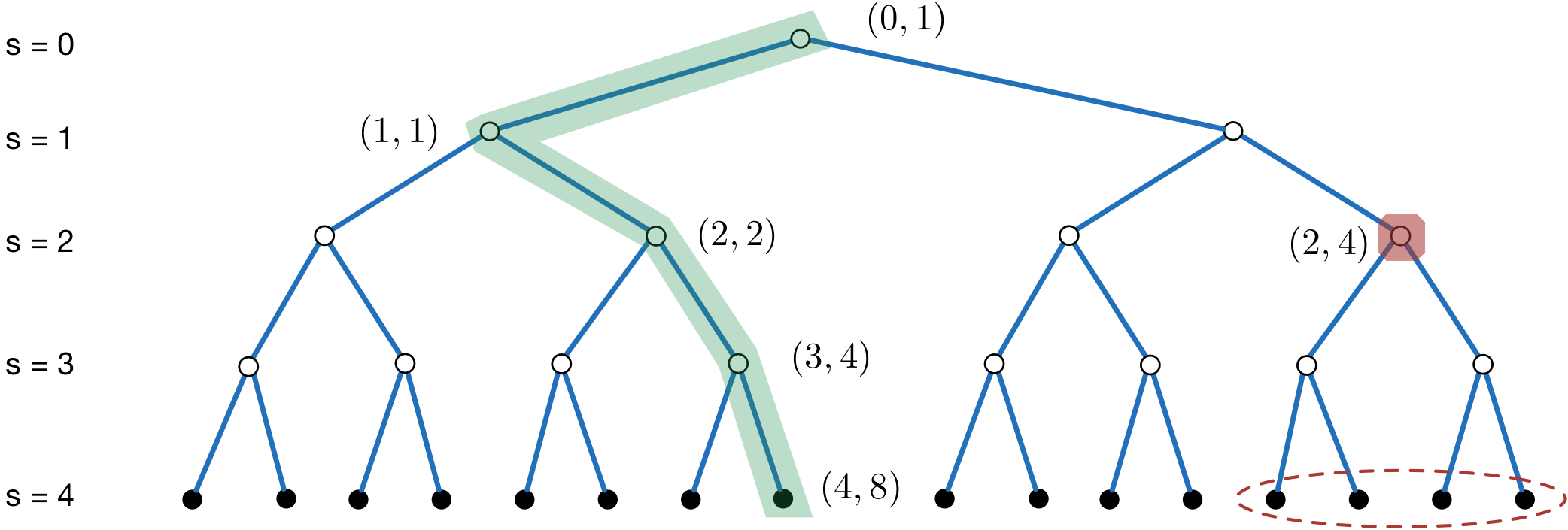}\\
\end{array}$}
\end{center}
 \vskip -0.10in
\caption{Illustration of $\mathcal{T}_{4}$ with $16$ leaf groups (solid dots). The green shaded region denotes a path set $\mathcal{A}_{8}$ from root node $(0,1)$ to leaf node $(4,8)$, through three intermediate nodes: $(1,1)$, $(2,2)$, and $(3,4)$. The descendant set $\mathcal{D}_{2,4}$ contains $4$ leaf nodes (located in the dashed circle) with a most recent common ancestor $(2,4)$ (indicated by the red octagon).  
}
\label{fig5:treepath}
\end{figure}

For $j=1,\hdots, m$, we denote by $\mathcal{A}_{j}$ the ``vertical" root-to-leaf  \emph{path set} in the tree $\mathcal{T}_{h}$ connected to the {\color{black}$j$-th} leaf group, $j=1,\hdots, m$. The path set $\mathcal{A}_{j}$ includes all the nodes that it visits from the root node to the leaf node $j$. Meanwhile, we denote the ``horizontal" \emph{descendant set} $\mathcal{D}_{s,\ell}$ as the set of leaf nodes who share a most recent common ancestor $(s,\ell)$, $(s,\ell) \in \mathcal{I}_{h-1}$, {\color{black}where $\mathcal{I}_{h-1}:=\{(s, \ell): 0 \leq s \leq h-1, 1\leq \ell \leq 2^{s}\}$ is the internal nodes in $\mathcal{T}_{h}$, in which each parent node $(s, \ell)$ has two child nodes $(s+1, 2\ell-2+t)$, $t\in\{1,2\}$}. Specifically, for $s=h$, we set $\mathcal{D}_{h,j}\equiv \mathcal{L}_{j}$, for $j = 1,\ldots, m$, for ease of notation.  Figure \ref{fig5:treepath} illustrates an example of a path set $\mathcal{A}_{8}$ and an example of a descendant set $\mathcal{D}_{2,4}$ with the common ancestor node $(2,4)$.

There exists a \emph{dual relationship} between random variables  and coefficients based on these two notations. 
We can attach a random variable $Z_{i,s,\ell}\in\mathbb{Z}$ to each node $(s,\ell)$, where $Z_{i,s,\ell} = N(\mathcal{E}_{i} \bigcap \mathcal{D}_{s, \ell})$ counts the appearances of POs in descendant set $\mathcal{D}_{s, \ell}$ and ${Z}_{i, s, \ell}=\sum_{j\in \mathcal{D}_{s, \ell}}{X}_{i,j}$. 
Letting $\gamma_{s, \ell}$ be the coefficient for node $( s, \ell)$ visited by the path $\mathcal{A}_{j}$,  we have ${\beta}_{j} = \sum_{(s, \ell)\in{A_{j}}}{\gamma}_{s, \ell}$.  Therefore, the sufficient reduction score $R_{\bd{\beta}}(\bd{x}_{i})$ introduced in Section \ref{sec3} can be re-expressed under the parameterization $\bd{\gamma}$ on  the partition tree $\mathcal{T}_{h}$, 
\eq{
R_{\bd{\beta}}(\bd{x}_{i}) = \bd{\beta}^{T}\bd{x}_{i}=\sum_{j=1}^{m}x_{i, j}\beta_{j}=\sum_{j=1}^{m}x_{i, j} \sum_{(s, \ell)\in{A_{j}}}{\gamma}_{s, \ell}=\sum_{s=0}^{h}\sum_{\ell =1}^{2^s}{z}_{i, s, \ell}\gamma_{s,\ell}=\bd{\gamma}^{T}\bd{z}_{i}= R_{\bd{\gamma}}(\bd{x}_{i}).
}
Clearly, $R_{\bd{\gamma}}(\bd{x}_{i})$ is also a linear sufficient reduction. The reparameterization changes neither the data likelihood of the PIR model \eqref{eq:model}, nor the sufficient reduction score. 

The reparameterization by the tree-structured PX scheme can be represented in matrix form as $\bd{\beta} = \bd{D}\bd{\gamma}$, where $\bd{D}$ is a $m \times L$ design matrix with binary entries, $m<L$. Each column in $\bd{D}$ can be interpreted as a basis function that encodes the piecewise smoothness at a different location and scale.  For example, assuming $h =3$, the number of leaves $m = 2^3 = 8$, $L = 15$, we have 
\setcounter{MaxMatrixCols}{20}
\eq{
\bd{D} = \begin{bmatrix}
    1     &  1  &  0 & 1  & 0 & 0  & 0 & 1 & 0 & 0  & 0 & 0 & 0 & 0 & 0\\
    1     &  1  & 0  & 1  & 0 & 0  & 0 & 0 & 1 & 0  & 0 & 0 & 0 & 0 & 0\\ 
    1     &  1  & 0  & 0  & 1 & 0  & 0 & 0 & 0 & 1  & 0 & 0 & 0 & 0 & 0\\ 
    1     &  1  & 0  & 0  & 1 & 0  & 0 & 0 & 0 & 0  & 1 & 0 & 0 & 0 & 0\\ 
    1     &  0  & 1  & 0  & 0 & 1  & 0 & 0 & 0 & 0  & 0 & 1 & 0 & 0 & 0\\ 
    1     &  0  & 1  & 0  & 0 & 1  & 0 & 0 & 0 & 0  & 0 & 0 & 1 & 0 & 0\\ 
    1     &  0  & 1  & 0  & 0 & 0  & 1 & 0 & 0 & 0  & 0 & 0  & 0 & 1 & 0\\ 
    1     &  0  & 1  & 0  & 0 & 0  & 1 & 0 & 0 & 0  & 0 & 0 & 0 & 0 & 1
\end{bmatrix}.
}

\subsection{Fused Generalized Double Pareto Prior}\label{sec: shrinkage}

{\color{black}We denote  by $\mathcal{C}_{s, \ell}$ the set of child nodes of $(s, \ell)$ and the set $\mathcal{F}_{s,\ell}$ all the nodes on the sub-branch rooted from node $(s,\ell)$. The tree $\mathcal{T}_{h}$ is originated from the root node $(0,1)$ with leaf nodes indexed by $\mathcal{L}:=\{(h, \ell): 1\leq \ell \leq 2^{h}\}$, $\mathcal{F}_{0,1}=\mathcal{I}_{h-1}\bigcup \mathcal{L}$.} In Section \ref{sec: operators}, we introduced two reductive operations: \emph{deletion} (if $\beta_{j}=0$) and \emph{fusion} (if $\beta_{j}=\beta_{j'}$, $\forall j, j'\in \mathcal{D}_{s, \ell}$)  on the leaf partitions along the tree. However, exhaustive search for all possible schemes is prohibitive. Even for a binary tree $\mathcal{T}_{4}$, these two operations in combination result in $458,330$ different schemes. Alternatively, effective execution of the following two operations can be induced by regularization in the $\bd{\gamma}$ parameterization:  
\be[(i)]
\item {\bf Deletion:} if ${\gamma}_{s, \ell}=0$, $\forall (s, \ell) \in \mathcal{A}_{j}$,  then $\beta_{j}=\sum_{(s, \ell)\in\mathcal{A}_{j}}{\gamma}_{s, \ell}=0$, which implies that the contributions of leaf predictor $j$  across all the scales are pruned out.
\item {\bf Fusion:} If $\forall (s',\ell)\in\mathcal{F}_{s, \ell}$, their child nodes satisfy ${\gamma}_{s'+1, 2\ell-1}={\gamma}_{s'+1, 2\ell}$, then $\beta_{j}=\beta_{j'}$, $\forall j, j'\in \mathcal{D}_{s, \ell}$, the leaf variables within $\mathcal{D}_{s, \ell}$ can be condensed into one variable. 
\ee
Note that the $\bd{\gamma}$ parameterization is redundant; both conditions above are sufficient but not necessary. Based on the above observations, we impose  generalized double Pareto (GDP) priors \citep{armagan2013generalized} on $\bd{\gamma}$ and the pairwise differences between sibling nodes, 
\eqn{
{\gamma}_{s, \ell} & \sim \textrm{GDP}(\xi_{1}, \alpha_{1}), \quad {\gamma}_{s'+1, 2\ell-1}-{\gamma}_{s'+1, 2\ell}\sim \textrm{GDP}(\xi_{2}, \alpha_{2}), \label{generalizedfusedGDP}
}
where $(s'+1, 2\ell-1), (s'+1, 2\ell)\in\mathcal{C}_{s', \ell}$,  $(s', \ell) \in \mathcal{I}_{h-1}$ and  $(s, \ell) \in \mathcal{F}_{0,1}$. The first prior encourages sparsity on the individual coefficients and the second prior promotes sparsity on the differences between pairs of siblings with a common parent node $(s', \ell)$. These priors lead to a generalized fused lasso-type penalty \citep{she2010sparse, tibshirani2011solution}, but the GDP prior corresponds to a reweighted $\ell_{1}$ penalty instead of $\ell_{1}$ (as will be seen in \eqref{eq: logsum}), which better approximates the $\ell_{0}$-like criterion \citep{candes2008enhancing}.  

For $\mathcal{T}_{h}$, the number of expanded parameters in $\bd{\gamma}$ is $L=2^{h+1}-1$. A natural question is whether there exists a multivariate prior on $\bd{\gamma}$ that could justify the compatibility of these two GDP priors. To see this, we use the latent variable representation of the GDP prior introduced in \cite{armagan2013generalized}. The first level of the hierarchy is written as, 
\eqn{
\gamma_{s, \ell} \sim \cN(0, \tau_{s, \ell}),\quad \gamma_{s'+1, 2\ell-1}-\gamma_{s'+1, 2\ell} \sim \cN(0, \phi_{s'+1, 2\ell-1, 2\ell}). \label{eq:firstGDP}
}

We postulate a multivariate normal prior for the $L$-dimensional vector $\bd{\gamma}\sim \mathcal{N}(\bd{0}, \bd{\Lambda}^{-1})$ with $L\times L$ precision matrix $\bd{\Lambda}$, whose log marginal density is different than that of priors in  \eqref{eq:firstGDP} only up to a constant. The entries in $\bd{\Lambda}$ as a function of $(\bd{\tau}, \bd{\phi})$ can be found by square completing. It takes a block-diagonal form as follows,  
\eqn{\bd{\Lambda}(\bd{\tau}, \bd{\phi}) = \mathrm{blockdiag}[  {1}/{\tau_{0,1}}; \bd{\Omega}_{0, 1};  \bd{\Omega}_{1, 1};  \bd{\Omega}_{1, 2}; \hdots;  \bd{\Omega}_{h-1, 1}, \hdots, \bd{\Omega}_{h-1, m/2}],\label{eq: blkd}
}  
where
\eq{
\bd{\Omega}_{s', \ell} = \begin{bmatrix}
          \frac{1}{\tau_{s'+1,2\ell-1}} & 0 \\
         0 & \frac{1}{\tau_{s'+1,2\ell}} \\
         \end{bmatrix} + \frac{1}{\phi_{s'+1, 2\ell-1, \ell}}\begin{bmatrix}
          1 & -1 \\
         -1& 1  \\
         \end{bmatrix}, \quad (s' ,\ell) \in \mathcal{I}_{h-1}. 
}
To complete the hierarchy of the multivariate GDP prior with $\bd{\gamma}\sim \mathcal{N}(\bd{0}, [\bd{\Lambda}(\bd{\tau}, \bd{\phi})]^{-1})$, we put $\tau_{s, \ell}\sim\mathrm{Exp}(\lambda_{s, \ell}^{2}/2)$, $\lambda_{s, \ell} \sim \mathrm{Ga}(\alpha_{1}, \eta_{1})$, $(s, \ell)\in\mathcal{F}_{0,1}$,  and 
$\phi_{s'+1, 2\ell-1, 2\ell} \sim\mathrm{Exp}(\nu_{s'+1, 2\ell-1, 2\ell}^{2}/2)$, $\nu_{s'+1, 2\ell-1, 2\ell} \sim \mathrm{Ga}(\alpha_{2}, \eta_{2})$,  $(s', \ell)\in\mathcal{I}_{h-1}$. So now we have a  fused generalized double Pareto (fGDP) prior, which promotes the desired form of structured sparsity in  $\bd{\gamma}$, and enjoys a latent variable representation that makes the parameter estimation straightforward. Integrating out the latent variables $\bd{\Psi}$,  we obtain the marginal density of the fGDP prior, denoted by $\textrm{fGDP}(\bd{\gamma}; \mathcal{T}_{h}, \alpha_{1}, \eta_{1}, \alpha_{2}, \eta_{2})$, whose logarithm takes the following form,  
\eqn{
\ln{p(\bd{\gamma})} &  = \sum_{(s,\ell)\in \mathcal{F}_{0,1}}\bigg[-\ln{(2\xi_{1})}-(\alpha_{1}+1)\ln{\bigg(1+\frac{|{\gamma}_{s,\ell}|}{\alpha_{1} \xi_{1}}}\bigg)\bigg]\cr 
& + \sum_{(s',\ell)\in\mathcal{I}_{h-1}}\bigg[-\ln{(2\xi_{2})}-(\alpha_{2}+1)\ln{\bigg(1+\frac{|{\gamma}_{s'+1, 2\ell-1}-{\gamma}_{s'+1,  2\ell}|}{\alpha_{2} \xi_{2}}}\bigg)\bigg],\label{eq: logsum}
}
where $\xi_{1}=\eta_{1}/\alpha_{1}$, $\xi_{2}=\eta_{2}/\alpha_{2}$. 
 
Importantly, through PX, we have transformed the problem of \emph{multiscale shrinkage} on the regression coefficients $\bd{\beta}$ across multiple scales on $\mathcal{T}_{h}$ into a \emph{structured shrinkage} problem on the expanded parameters $\bd{\gamma}$, which can be conveniently addressed via the proposed fGDP prior.  The hierarchical-Bayes representation of the multiscale shrinkage prior on $\bd{\beta}$ can be obtained via integrating out $\bd{\gamma}$; we have the conditional prior $\bd{\beta}|\bd{\tau}, \bd{\phi}\sim \mathcal{N}(\bd{0}, \bd{D}\bd{\Lambda}(\bd{\tau}, \bd{\phi})^{-1}\bd{D}^{T})$ and the priors on the latent variables $\{\bd{\tau}, \bd{\phi}, \bd{\lambda}, \bd{\nu}\}$ do not change. However, the precision matrix $\bd{D}\bd{\Lambda}(\bd{\tau}, \bd{\phi})^{-1}\bd{D}^{T}$ no longer exhibits a sparse block-diagonal structure as in \eqref{eq: blkd}, and {\color{black}the resulting EM procedure of estimating $\bd{\beta}$ is less tractable than estimating $\bd{\gamma}$ as the former involves intractable expectations.}

\section{Parameter Estimation}\label{sec5}
We further assume each replicate is collected within a time window of length $t_{i}$ (known), $i = 1, \hdots, n$.  To accommodate potential overdispersion and dependencies, we incorporate random effects in the model. The Poisson log-linear mixed regression model is written as follows, 
\eqn{
x_{i, j} \sim \textrm{Poisson}(\mu_{i, j}), \quad \mu_{i, j} = t_{i}e^{\eta_{i, j}},\quad \eta_{i, j} = a+ b_{i}+c_{j}+y_{i}\beta_{j},\label{eq: plmm}
}
where $(\beta_{1}, \hdots, \beta_{m})$ is the fixed effect slope parameter for the $m$ \textit{simple Poisson mixed regression} model \citep{hall2011asymptotic}. The fixed effects measure the common association between the predictors and response, while the random effects allow replicates or leaf groups to have their own baseline rates. The total, column and row random effects are  $a$, $\bd{b}$, and $\bd{c}$, respectively.  Constraints are needed for the identifiability of row and column scores $b_{i}$ and $c_{j}$, so we use the corner constraint \citep{yee2003reduced} $b_{1} \equiv c_{1} \equiv 0$ in this article. Gaussian priors on $a$, $\bd{b} = [b_{2}, \hdots, b_{n}]$ and $\bd{c} = [c_{2}, \hdots, c_{m}]$ are specified as follows, 
\eq{
a \sim \cN(0, \omega_{a}), \quad b_{i}\sim \cN(0, \omega_{b}), \quad c_{j}\sim \cN(0, \omega_{c}), \quad i = 2,\hdots, n, \quad j = 2, \hdots, m, 
}
with unknown variance parameters $\bd{\omega} = [\omega_{a}, \omega_{b}, \omega_{c}]$. Since $\beta_{j} = \bd{d}_{j}^{T}\bd{\gamma}$, we have $\eta_{i,j} = a+b_{i}+c_{j}+y_{i}\bd{d}_{j}^{T}\bd{\gamma}$. Note that the sufficiency of $R_{\bd{\gamma}}(\bd{x}_{i})$ established in Section \ref{subsec:PIR} still holds conditional on the random effect terms \citep{taddy2013multinomial}. In the next section, we introduce a penalized likelihood estimator of $\bd{\gamma}$,  under the conditional likelihood ${\ell}(\bd{\gamma}, \bd{\omega}):=\ln{p(\bd{X}|\bd{y}, \bd{\gamma}, \bd{\omega})}$ with the proposed fGDP prior $\bd{\gamma}\sim\textrm{fGDP}(\alpha_{1}, \eta_{1}, \alpha_{2}, \eta_{2})$ guided by $\mathcal{T}_{h}$. 

\subsection{{\color{black}Variational Expectation Maximization (EM)}}
The hierarchical-Bayes representation of the fGDP prior facilitates an iterative EM-type algorithm for penalized estimation with \eqref{eq: logsum}. We adopt the type-I estimation framework \citep{figueiredo2003adaptive}, which treats {\color{black}$\bd{\Psi}:=\{\bd{\tau}, \bd{\phi}, \bd{\lambda}, \bd{\nu}\}$}  as latent variables and $\{\bd{\gamma}, \bd{\omega}\}$ as parameters to optimize. The conditional likelihood of the model in \eqref{eq: plmm} involves a $(n+m-1)$-dimensional integral, 
\eq{
\ln{p(\bd{X}|\bd{y}, \bd{\gamma}, \bd{\omega})} = \ln{\int p(\bd{X}|\bd{y}, \bd{\gamma}, {a}, \bd{b}, \bd{c})p(a|\omega_{a})p(\bd{b}|\omega_{b})p(\bd{c}|\omega_{c}) \textsf{d} a \textsf{d} \bd{b} \textsf{d} \bd{c}},   
}
which is nonanalytic. Alternatively, we take a Gaussian variational approximation (GVA) of the posteriors of random effect variables $\bd{U}:=\{a, \bd{b}, \bd{c}\}$, which provides a lower bound $\underline{\ell}(\bd{\gamma}, \bd{\omega}, \bd{\zeta}, \bd{\kappa})$ of  ${\ell}(\bd{\gamma}, \bd{\omega})$. Statistical properties of GVA for generalized linear mixed models are studied in \cite{hall2011theory, hall2011asymptotic} and \cite{ormerod2012gaussian}, from a likelihood-based perspective. The resulting GVA estimator differs from the MLE but is asymptotically valid. 

In our setting, the alternating steps are guaranteed to increase the following objective function \citep{neal1998view}, 
\eq{
\mathcal{F}(q, \bd{\gamma}, \bd{\omega}) = \langle \log{P(\bd{X}; \bd{\gamma}, \bd{\Psi}, \bd{U}|\bd{y}, \bd{\omega})}\rangle_{q(\bd{\Psi}, \bd{U})} + H[q(\bd{\Psi}, \bd{U})], 
}
where $q(\bd{\Psi}, \bd{U}) = q(\bd{\Psi})q_{\bd{\zeta}, \bd{\kappa}}(\bd{U})$ naturally decouples into a factorized form, in which $q(\bd{\Psi})$ is left in free-form and $q_{\bd{\zeta}, \bd{\kappa}}(\bd{U})$ is parameterized as Gaussian with diagonal covariance. With $t$ indexing the iterations, the overall algorithm contains the following alternating steps: 
\bi
\item {\bf E-step:} Optimize $\mathcal{F}(q, \bd{\gamma}, \bd{\omega})$ w.r.t. the distribution of latent variables $q(\bd{\Psi})$
\eq{
q^{(t)}(\bd{\Psi}):=\argmax_{q(\bd{\Psi})}\mathcal{F}(q(\bd{\Psi}), q^{(t-1)}(\bd{U}), \bd{\gamma}^{(t-1)}, \bd{\omega}^{(t-1)}). 
}
\item {\bf Variational E-step:} Update the Gaussian variational parameters $\{\bd{\zeta}, \bd{\kappa}\}$ such that 
\eq{
\mathcal{F}(q^{(t)}(\bd{\Psi}), q^{(t)}(\bd{U}), \bd{\gamma}^{(t-1)}, \bd{\omega}^{(t-1)})\geq \mathcal{F}(q^{(t)}(\bd{\Psi}), q^{(t-1)}(\bd{U}), \bd{\gamma}^{(t-1)}, \bd{\omega}^{(t-1)}). 
}
\item {\bf M-step.} Update the model parameters $\{\bd{\gamma}, \bd{\omega}\}$ such that 
\eq{
\mathcal{F}(q^{(t)}(\bd{\Psi}), q^{(t)}(\bd{U}), \bd{\gamma}^{(t)}, \bd{\omega}^{(t)})\geq \mathcal{F}(q^{(t)}(\bd{\Psi}), q^{(t)}(\bd{U}), \bd{\gamma}^{(t-1)}, \bd{\omega}^{(t-1)}). 
}
\ei
In this algorithm, the variational parameters $\{\bd{\zeta}, \bd{\kappa}\}$ and the model parameters $\{\bd{\gamma}, \bd{\omega}\}$ are updated with gradient updates instead of exact maximization \citep{lange1995gradient, lange1995quasi}. Therefore, it is a generalized EM algorithm \citep{dempster1977maximum, neal1998view}, as both the E-step and M-step are taken partially. Note that the latent variables $\bd{\Psi}$ only appear in the prior, and the random effect terms $\bd{U}$ only appear in the likelihood, 
\eq{
\mathcal{F}(q, \bd{\gamma}, \bd{\omega})  =  \langle \ell_{1}(\bd{\Psi}; \bd{\gamma})\rangle_{q(\bd{\Psi})} + \langle \ell_{2}(\bd{U}; \bd{\gamma}, \bd{\omega})\rangle_{q(\bd{U})},  
}
so we can discuss them separately. 

\subsection{Closed-Form Expectations in the Shrinkage Prior}
We compute the expected value w.r.t $\bd{\Psi}$ in the complete log-posterior, given the current parameter estimates and the observed data. Note that the entropy term does not depend on $(\bd{\gamma}, \bd{\omega})$, so the relevant term in the E-step is
\eq{
\langle \ell_{1}(\bd{\Psi}; \bd{\gamma})\rangle_{q(\bd{\Psi})}  = \mathbb{E}_{p(\bd{\Psi}|{\bd{\gamma}}^{(t)}, {\bd{\omega}}^{(t)}, \bd{X}, \bd{y})}[\ell_{1}(\bd{\Psi}, \bd{\gamma})],  
}
where 
\eq{
\ell_{1}(\bd{\Psi}, \bd{\gamma}) & =  \sum_{(s,\ell)\in \mathcal{F}_{0,1}}\ln p(\gamma_{s,\ell}, \tau_{s,\ell}, \lambda_{s,\ell}|\alpha_{1}, \eta_{1})\cr   
&+\sum_{(s',\ell)\in\mathcal{I}_{h-1}}\bigg[\ln p(\delta_{s'+1, 2\ell-1 ,2\ell}, \phi_{s'+1, 2\ell-1 ,2\ell}, \nu_{s'+1, 2\ell-1 ,2\ell}|\alpha_{2}, \eta_{2})\bigg].
}
According to the Gaussian scale mixture (GSM) representation of the GDP prior, 
\eq{
\ln p(\gamma_{s,\ell}, \tau_{s,\ell}, \lambda_{s,\ell}|\alpha_{1}, \eta_{1}) = \ln p(\gamma_{s,\ell}|\tau_{s,\ell})+\ln p(\tau_{s,\ell}|\lambda_{s,\ell})+\ln p(\lambda_{s,\ell}|\alpha_{1}, \eta_{1}),} and denoting the pairwise differences as $\delta_{r, 2\ell-1 ,2\ell} : = \gamma_{r, 2\ell-1}-\gamma_{r, 2\ell}$, $r = 1, \hdots, h$, we have 
\eq{
\ln p(\delta_{r, 2\ell-1 ,2\ell}, \phi_{r, 2\ell-1 ,2\ell}, \nu_{r, 2\ell-1 ,2\ell}|\alpha_{2}, \eta_{2}) & = \ln p(\delta_{r, 2\ell-1 ,2\ell}|\phi_{r, 2\ell-1 ,2\ell})+\ln p(\phi_{r, 2\ell-1 ,2\ell}|\nu_{r, 2\ell-1 ,2\ell})\cr & +\ln p(\nu_{r, 2\ell-1 ,2\ell}|\alpha_{2}, \eta_{2}).} 

Given the estimates from the previous iteration ${\bd{\gamma}}^{(t)}$, the conditional posterior of latent variables $(\bd{\tau}, \bd{\lambda})$ factorizes as
\eq{
p(\bd{\tau}, \bd{\lambda}|-) = \prod_{l\in \mathcal{I}_{h-1}\bigcup \mathcal{L}}p({\tau}_{l}, {\lambda}_{l}|-), \quad p({\tau}_{l}, {\lambda}_{l}|-)=p({\tau}_{l}| {\lambda}_{l}, -)p(\lambda_{l}|-),}
 where $({\tau}_{l}| {\lambda}_{l}, -)\sim \mathrm{GIG}(0.5,  \lambda_{l}^{2}, \gamma_{l}^{2})$.  
Integrating out $\tau_{l}$, we have $(\gamma_{l}|\lambda_{l})\sim \mathrm{DE}(\gamma_{l}; 0, 1/\lambda_{l})$ and $(\lambda_{l}|\alpha_{1}, \eta_{1})\sim \mathrm{Ga}(\lambda_{l}; \alpha_{1}, \eta_{1})$, so $(\lambda_{l}|\gamma_{l}, \alpha_{1}, \eta_{1})\sim \mathrm{Ga}(\alpha_{1}+1, {|\gamma_{l}|}+\eta_{1})$, where $\mathrm{DE}(x; \mu=0, b)=\exp{(-|x|/b)}/{2b}$ refers to the Laplace distribution with scale parameter $b = 1/\lambda$ and $\textrm{GIG}(x;a,b,p)$ denotes the Generalized Inverse Gaussian (GIG) distribution, $\textrm{GIG}(x;a,b,p)={0.5(a/b)^{{p}/{2}}}x^{p-1}\exp\left(-(ax+{b}/{x})/2\right)/K_{p}(\sqrt{ab})$, $(x>0)$, and $K_{p}(\theta)$ is the modified Bessel function of the second kind.

Similarly, given the estimates of the ${\bd{\delta}}^{(t)}$, the conditional posterior of latent variables $(\bd{\phi}, \bd{\nu})$ also factorizes. Thus for every $(u,w)\in \mathcal{C}_{s', \ell}$, $(s',\ell)\in\mathcal{I}_{h-1}$, we have $({\phi}_{u,w}| {\nu}_{u,w}, -)\sim \mathrm{GIG}(0.5, \nu_{u,w}^{2}, \delta_{u,w}^{2})$, 
and integrating out $\phi_{u,w}$, we have $(\delta_{u,w}|\nu_{u,w})\sim \mathrm{DE}(\delta_{u,w}; 0, 1/\nu_{u,w})$ and $(\nu_{u,w}|\alpha_{2}, \eta_{2})\sim \mathrm{Ga}(\nu_{u,w}; \alpha_{2}, \eta_{2})$, so $(\nu_{u,w}|\delta_{u,w}, \alpha_{2}, \eta_{2})\sim \mathrm{Ga}(\alpha_{2}+1, {|\delta_{u,w}|}+\eta_{2})$. Therefore, 
\eqn{
\langle \ell_{1}(\bd{\Psi}; \bd{\gamma})\rangle_{q(\bd{\Psi})}  =  \langle \ell_{1}(\bd{\Psi}; \bd{\gamma})\rangle_{p(\bd{\Psi}|\bd{\gamma}^{(t)}, -)}  =  -\sum_{l=1}^{L}\frac{\gamma_{l}^{2}}{2}\langle{\tau_{l}^{-1}}\rangle -\sum_{(u,w)\in \mathcal{C}_{s',\ell}}\frac{\delta_{u,w}^{2}}{2}\langle{\phi_{u,w}^{-1}}\rangle. \label{eq:qfunction}
}
We only need to find $\langle{\tau_{l}}^{-1}\rangle:=\langle{\rho_{l}}\rangle$. According to the change of variable formula, $f(\rho_{l})=\mathrm{GIG}(\tau_{l}^{-1}; p, a, b)\rho_{l}^{-1}=\mathrm{GIG}(\rho_{l}; -0.5, b, a)=\mathrm{InvGau}(\rho_{l}; \sqrt{{\lambda_{l}^{2}}/{{\gamma_{l}}^{2(t)}}}, \lambda_{l}^{2})$, 
we have
\eqn{
\mathbb{E}_{p(\rho_{l}|\lambda_{l}, -)}[{\rho_{l}}] & = {\lambda_{l}}/{|{\gamma_{l}}^{(t)}|}, \cr \langle{\rho_{l}}\rangle & = \mathbb{E}_{p(\lambda_{l}| -)}\bigg[\mathbb{E}_{p(\rho_{l}|\lambda_{l}, -)}({\rho_{l}})\bigg] = \frac{1}{|{\gamma_{l}}^{(t)}|}\mathbb{E}_{p(\lambda_{l}| -)}[\lambda_{l}]=\frac{(\alpha_{1}+1)}{|{\gamma_{l}}^{(t)}|[{|{\gamma_{l}}^{(t)}|}+\eta_{1}]},\label{eq:shrinkage1}
}
and similarly denoting $\langle{\phi_{u,w}}^{-1}\rangle:=\langle{\upsilon_{u,w}}\rangle$, we obtain
\eqn{
\langle{\upsilon_{u,w}}\rangle  =\langle{\phi_{u,w}^{-1}}\rangle  =\frac{(\alpha_{2}+1)}{|{\delta_{u,w}}^{(t)}|[{|{\delta_{u,w}}^{(t)}|}+\eta_{2}]}.\label{eq:shrinkage2}
}

The GSM representation of the GDP priors determines a reweighting rule, as shown in \eqref{eq:shrinkage1} and \eqref{eq:shrinkage2}, in which the weights depend only on the current estimate of the parameter $\bd{\gamma}^{(t)}$ (or its differences), and the hyperparameters $(\alpha_{1}, \eta_{1}, \alpha_{2}, \eta_{2})$.

The penalty terms in \eqref{eq:qfunction} can be organized in a quadratic form, where the block-diagonal matrix $\wt{\bd{\Lambda}}$ can again be  found by square completing, 
\eq{\wt{\bd{\Lambda}}= \mathrm{blockdiag}[\langle{\rho_{0,1}}\rangle; \wt{\bd{\Omega}}_{0, 1};  \wt{\bd{\Omega}}_{1, 1};  \wt{\bd{\Omega}}_{1, 2}; \hdots;  \wt{\bd{\Omega}}_{h-1, 1}, \hdots, \wt{\bd{\Omega}}_{h-1, m/2}],
}  
where
\eq{
\wt{\bd{\Omega}}_{s', \ell} = \begin{bmatrix}
          \langle{\rho_{s'+1,2\ell-1}}\rangle & 0 \\
         0 &  \langle{\rho_{s'+1,2\ell}}\rangle\\
         \end{bmatrix} +  \langle{\upsilon_{s'+1, 2\ell-1, \ell}}\rangle\begin{bmatrix}
          1 & -1 \\
         -1& 1  \\
         \end{bmatrix}, \quad (s' ,\ell) \in \mathcal{I}_{h-1}. 
}
Therefore, $\langle \ell_{1}(\bd{\Psi}; \bd{\gamma})\rangle_{q(\bd{\Psi})}  =  \langle \ell_{1}(\bd{\Psi}; \bd{\gamma})\rangle_{p(\bd{\Psi}|\bd{\gamma}^{(t)}, -)}  = -\bd{\gamma}^{T}\wt{\bd{\Lambda}}\bd{\gamma}/2$,
which is the structured penalty that favors models with simpler structures conforming to $\mathcal{T}_{h}$. The quadratic form is concave and differentiable, which makes gradient-based optimization methods suitable.

\subsection{Gaussian Variational Approximation of the Likelihood}
In our case, the log-likelihood for the Poisson mixed model in \eqref{eq: plmm} is 
\eq{
\ln{p(\bd{X}|\bd{y}, \bd{\gamma}, {a}, \bd{b}, \bd{c})}  = \sum_{i=1}^{n} \sum_{j=1}^{m}\bigg[x_{i,j}{(\ln{t_{i}}+\eta_{i,j})}-t_{i}\exp{(\eta_{i,j})}-\ln{(x_{i,j}!)}\bigg]. 
}
The log-priors for the random effect terms are
\eq{
\ln{p(a|\omega_{a})}  = -\frac{1}{2}\ln{(2\pi \omega_{a})}-\frac{a^{2}}{2\omega_{a}},
} 
\eq{
\ln{p(\bd{b}|\omega_{b})} = \sum_{i=2}^{n}\ln{p(b_{i}|\omega_{b})} = -\frac{(n-1)}{2}\ln{(2\pi \omega_{b})}-\sum_{i=2}^{n}\frac{b_{i}^{2}}{2\omega_{b}},
}
\eq{
\ln{p(\bd{c}|\omega_{c})} = \sum_{j=2}^{m}\ln{p(b_{j}|\omega_{c})} = -\frac{(m-1)}{2}\ln{(2\pi \omega_{c})}-\sum_{j=2}^{m}\frac{c_{j}^{2}}{2\omega_{c}}.
}
Let $q({a}) = \cN(\zeta^{a}, \kappa^{a})$, $q(b_{i}) = \cN(\zeta^{b}_{i}, \kappa^{b}_{i})$,  $q(c_{j}) = \cN(\zeta^{c}_{j}, \kappa^{c}_{j})$\footnote{Note that for $b_{1}$ and $c_{1}$, we have fixed their value to $0$ therefore for notational convenience, we assume $\zeta^{b}_{1} =\zeta^{c}_{1} = 0$ and $\kappa^{b}_{1} =\kappa^{c}_{1} = 0$ in the likelihood term.
}, where $\{\zeta^{a}, \bd{\zeta}^{b}, \bd{\zeta}^{c}\}$ are the mean parameters and  $\{\kappa^{a}, \bd{\kappa}^{b}, \bd{\kappa}^{c}\}$ are all positive parameters for the variances.  Assuming the variational proposals are independent,  the lower bound of $\ln{p(\bd{X}|\bd{y}, \bd{\gamma}, \bd{\omega})}$ is 
\eq{
 \underline{\ell}(\bd{\gamma}, \bd{\omega}, \bd{\zeta}, \bd{\kappa})  & = \mathbb{E}_{q}\bigg[\sum_{i=1}^{n}\sum_{j=1}^{m}\ln{p({x}_{i, j}|{y}_{i}, a, b_{i}, c_{j},  \bd{\gamma})} \bigg] +\mathbb{E}_{q(a)}\bigg[\ln{p(a|\omega_{a})}-\ln{q(a)}\bigg]  \cr & +\sum_{i=2}^{n}\mathbb{E}_{q(b_{i})}\bigg[\ln{p(b_{i}|\omega_{b})}-\ln{q(b_{i})}\bigg]+\sum_{j=2}^{m}\mathbb{E}_{q(c_{j})}\bigg[\ln{p(c_{j}|\omega_{c})}-\ln{q(c_{j})}\bigg]. 
}
Denoting $\underline{\ell}(\bd{\gamma}, \bd{\omega}, \bd{\zeta}, \bd{\kappa})  = \langle \ell_{2}(\bd{U}; \bd{\gamma}, \bd{\omega})\rangle_{q(\bd{U})}$, we have
\eq{
\langle \ell_{2}(\bd{U}; \bd{\gamma}, \bd{\omega})\rangle_{q(\bd{U})} & = \sum_{i=1}^{n} \sum_{j=1}^{m}x_{i,j}{\bigg(\zeta^{a}+\zeta^{b}_{i}+\zeta^{c}_{j}+y_{i}\bd{d}_{j}^{T}\bd{\gamma}\bigg)} \cr & 
- \sum_{i=1}^{n} \sum_{j=1}^{m}t_{i}\exp{\bigg(\zeta^{a}+\zeta^{b}_{i}+\zeta^{c}_{j}+\frac{1}{2}(\kappa^{a}+\kappa^{b}_{i}+\kappa^{c}_{j})+y_{i}\bd{d}_{j}^{T}\bd{\gamma}\bigg)}\cr 
&-\frac{1}{2\omega_{a}}{\bigg[(\zeta^{a})^{2}+\kappa^{a}\bigg]} -\frac{1}{2\omega_{b}}\sum_{i=2}^{n}{\bigg[(\zeta^{b}_{i})^{2}+\kappa^{b}_{i}\bigg]} -\frac{1}{2\omega_{c}}\sum_{j=2}^{m}{\bigg[(\zeta^{c}_{j})^{2}+\kappa^{c}_{j}\bigg]}\cr & -\frac{1}{2}\ln{(\omega_{a})} -\frac{n-1}{2}\ln{(\omega_{b})} -\frac{m-1}{2}\ln{(\omega_{c})} \cr & +\frac{1}{2}\ln{(\kappa^{a})}+\frac{1}{2}\sum_{i=2}^{n}\ln{(\kappa^{b}_{i})}+\frac{1}{2}\sum_{j=2}^{m}\ln{(\kappa^{c}_{j})}+\frac{n+m-1}{2}.
}
So in the variational E-step and the M-step, we update  the variational parameters $\{\bd{\zeta}, \bd{\kappa}\}$ and the model parameters $\bd{\gamma}$ through a quasi-Newton method with objective function $Q(\bd{\gamma}, \bd{\omega}, \bd{\zeta}, \bd{\kappa}):=\underline{\ell}(\bd{\gamma}, \bd{\omega}, \bd{\zeta}, \bd{\kappa}) 
   -\bd{\gamma}^{T}\wt{\bd{\Lambda}}\bd{\gamma}/2$, which only requires us to specify the first-order gradients (detailed in Appendix \ref{appendix: a}). 

In each M-step, we can also choose to optimize the prior parameter $\bd{\omega}$ via a fixed-point update. Setting the gradients $D_{\omega_{a}}{Q}  = D_{\omega_{b}}{Q} = D_{\omega_{c}}{Q} = 0$, we obtain, 
\eq{
\omega_{a} =\bigg((\zeta^{a})^{2}+\kappa^{a}\bigg), \quad \omega_{b} = \frac{1}{n-1}\sum_{i=1}^{n}\bigg((\zeta_{i}^{b})^{2}+\kappa_{i}^{b}\bigg),  \quad  \omega_{c} = \frac{1}{m-1}\sum_{j=1}^{m}\bigg((\zeta_{j}^{c})^{2}+\kappa_{j}^{c}\bigg). 
}
As suggested in \cite{armagan2013generalized}, the hyper-parameters $\{\alpha_{1}, \eta_{1}, \alpha_{2}, \eta_{2}\}$ can be either fixed or pre-learned from an initial Bayesian analysis based on griddy Gibbs sampling \citep{ritter1992facilitating}. 

\subsection{Computational Complexity}

Our spinlets method is scalable to handle millions of POs. {\color{black}The complexity analysis and the running time of the recursive graph partitioning algorithm are detailed in Appendix \ref{appendix1}.} In balancing the per-iteration cost with the convergence rate in the variational EM algorithm, we adopt a quasi-Newton method with the L-BFGS algorithm \citep{liu1989limited}, which uses a predetermined $c_{0}=100$ number of previous steps to form a low-rank Hessian approximation with complexity $O(mc_{0})$. As will be illustrated in Figure \ref{fig6: convergence}, the variational EM algorithm converges very fast in practice. We measure the CPU time of these procedures on a standard laptop computer (Macbook Air, 1.6 GHz Intel Core i5, 8 GB 1600 MHz DDR3, Intel HD Graphics 6000 1536 MB). For the StatsBomb World Cup data with $h=9$ and $n = 128$, each iteration of the variational EM algorithm takes about $14.91$ seconds to run.

\section{Applications}\label{sec6}
In this section, we first compare the performance of various penalties within the GDP and fGDP families in a number of simulated examples (Section \ref{sec6a}), and then use spinlets as an exploratory tool for visualizing spatial interaction networks from the FIFA World Cup 2018 dataset, while using the supervised information of interest (Section \ref{sec6b}).    

\subsection{Simulation Study}\label{sec6a}

We generate $n=\{25, 50, 100, 200\}$ observations from the model $x_{i, j} \sim \textrm{Poisson}(\mu_{i, j})$, $\mu_{i, j} = t_{i}e^{\eta_{i, j}}$, $\eta_{i, j} = a+ b_{i}+c_{j}+y_{i}\beta_{j}^{*}$, $t_{i}\sim\textrm{Ga}(2,1)$, $y_{i}\sim\textrm{Poisson}(0.5)$, $i = 1,\hdots, n$, $j = 1, \hdots, m$, and $a \sim \cN(0, 0.1)$, $b_{i'}\sim \cN(0, 0.1)$, $c_{j'}\sim \cN(0, 0.1)$, for $i' = 2,\hdots, n$, $j' = 2, \hdots, m$, $m = 2^{h}$, $h = 5$. We assume $\bd{\beta}^{*}$ to be signals with multiscale structures in the following configurations: 
{
\be[(a)]
\item $\bd{\beta}^{*} = (1,1,0,0,1,1,0,0, 1,1,-1,-1,0,0,-1,-1, 1,1,0,0,1,1,0,0,-1,-1,1,1,0,0,1,1)$,
\item $\bd{\beta}^{*} = (\underbrace{1,\hdots,1}_{4}, \underbrace{0,\hdots,0}_{4},\underbrace{-1,\hdots,-1}_{4},\underbrace{0,\hdots,0}_{4}, \underbrace{1,\hdots,1}_{4},\underbrace{-1,\hdots,-1}_{4},\underbrace{0,\hdots,0}_{4},\underbrace{1,\hdots,1}_{4})$,
\item $\bd{\beta}^{*} = (\underbrace{1,\hdots,1}_{8}, \underbrace{0,\hdots,0}_{8},\underbrace{-1,\hdots,-1}_{8},\underbrace{0,\hdots,0}_{8})
$,
\item $\bd{\beta}^{*} = (1,1,0,0,\underbrace{-1,\hdots,-1}_{4},, \underbrace{0,\hdots,0}_{8}, \underbrace{1,\hdots,1}_{16})
$.
\ee
}
\subsubsection{Baseline Methods}
Denoting the GDP prior as $\textrm{GDP}(\alpha, \eta)$, we consider the following baseline methods (detailed derivations provided in Appendix \ref{appendix: b}) operating on the original parameter $\bd{\beta}$: 
\newcounter{baselines}
{
\be[(1)]
\item GDP-0: no prior, $\beta_{j} \sim \textrm{GDP}(-1, 1)$, and $j = 1, \hdots, m$,
\item GDP prior with default parameters, $\beta_{j} \sim \textrm{GDP}(1, 1)$, and $j = 1, \hdots, m$,  
\item Fused lasso signal approximation (FLSA) \citep{friedman2007pathwise} implemented with the GDP prior: $\beta_{j} \sim \textrm{GDP}(1, 1)$, $\beta_{j'}-\beta_{j'+1} \sim \textrm{GDP}(1, 1)$, and $j' = 1,\hdots, m-1$,
\item Pairwise fused lasso (PFL-S) \citep{petry2011pairwise} with the deletion and fusion penalties  weighted by $0.8$ and $0.2$ respectively,   implemented with the GDP prior: $\beta_{j}\sim \textrm{GDP}(1, 1)$, and $\beta_{j}-\beta_{k} \sim \textrm{GDP}(1,1)$, $j, k = 1,\hdots, m$, and $j\neq k$. 
\item Pairwise fused lasso (PFL-F) \citep{petry2011pairwise} with the deletion and fusion penalties  weighted by $0.2$ and $0.8$ respectively (implementation is similar as above).
\setcounter{baselines}{\value{enumi}}
\ee
}
 
We also consider the following structural regularizers on the expanded parameter $\bd{\gamma}$. All of them are derived from the $\textrm{fGDP}(\alpha_{1}, \eta_{1}, \alpha_{2}, \eta_{2})$ family: 
{
\be[(1)]
\setcounter{enumi}{\value{baselines}}
\item fGDP-S: deletion only, $\bd{\gamma}\sim\textrm{fGDP}(1, 1, -1, 1)$,
\item fGDP-F: fusion only, $\bd{\gamma}\sim\textrm{fGDP}(-1, 1, 1, 1)$,
\item fGDP: with default parameters, $\bd{\gamma}\sim\textrm{fGDP}(1, 1, 1, 1)$,
\item fGDP-NJ: with Normal-Jeffrey's parameters, $\bd{\gamma}\sim\textrm{fGDP}(0, 0, 0, 0)$.
\setcounter{baselines}{\value{enumi}}
\ee}
To investigate the sensitivity to the hyperparameters $\{\alpha_{1}, \eta_{1}, \alpha_{2}, \eta_{2}\}$, we additionally include: (10) fGDP1: $\bd{\gamma}\sim\textrm{fGDP}(1, 0.1, 1, 0.1)$, (11) fGDP2: $\bd{\gamma}\sim\textrm{fGDP}(1, 0.01, 1, 0.01)$, (12) fGDP3: $\bd{\gamma}\sim\textrm{fGDP}(1, 0.001, 1, 0.001)$, (13)  fGDP4: $\bd{\gamma}\sim\textrm{fGDP}(0.5, 0.01, 0.5, 0.01)$, (14) fGDP5: $\bd{\gamma}\sim\textrm{fGDP}(2, 0.01, 2, 0.01)$, and (15)  fGDP6: $\bd{\gamma}\sim\textrm{fGDP}(5, 0.01, 5, 0.01)$. All these $15$ methods share the same random initialization of parameters.  {\color{black}The TFL penalties \citep{wang2017constructing} are not considered, as they only deal with tree-guided variable fusion and are limited to linear models in a completely different forward regression setting.}

\begin{figure}[hbpt] 
\vskip -0.00in
\begin{center}
\footnotesize{
$\begin{array}{ccc}
\hspace{-0.3cm}\includegraphics[height=0.33\textwidth, width=0.33\textwidth]{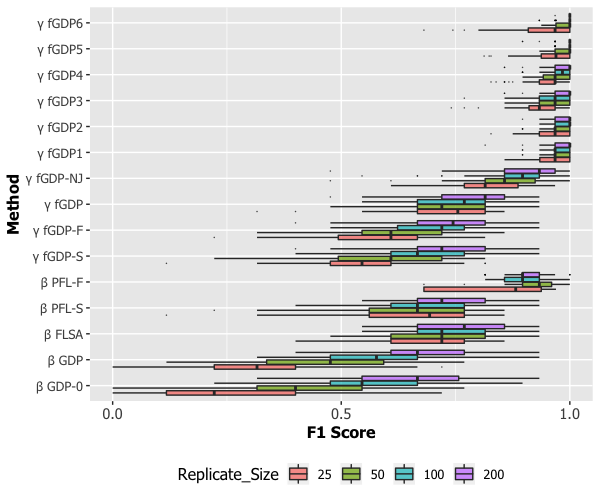} & \hspace{-0.3cm}\includegraphics[height=0.33\textwidth, width=0.33\textwidth]{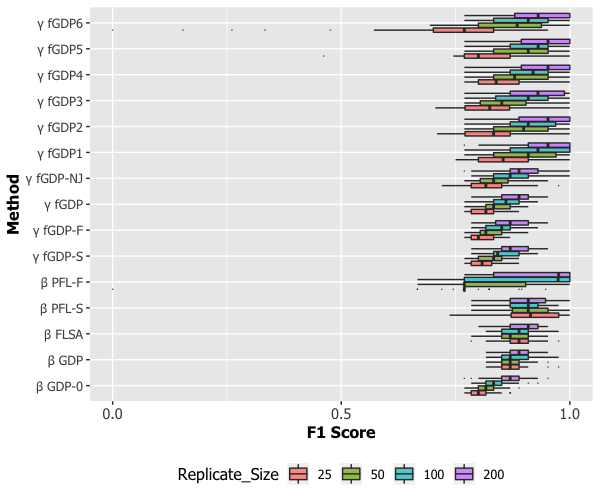}& \hspace{-0.3cm}\includegraphics[height=0.33\textwidth, width=0.33\textwidth]{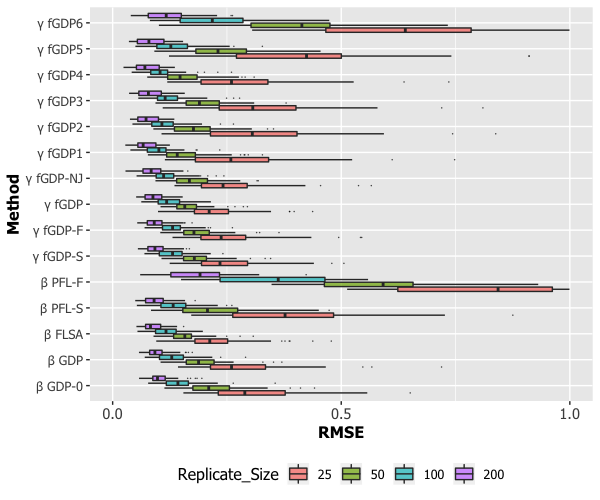}\\[-0.6cm]
\hspace{-0.3cm}\includegraphics[height=0.33\textwidth, width=0.33\textwidth]{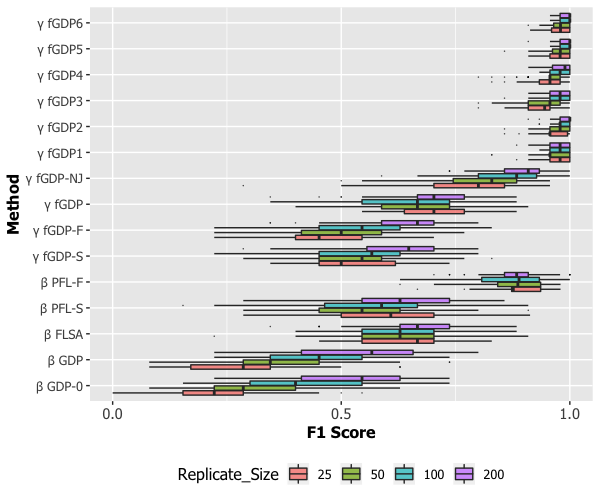} & \hspace{-0.3cm}\includegraphics[height=0.33\textwidth, width=0.33\textwidth]{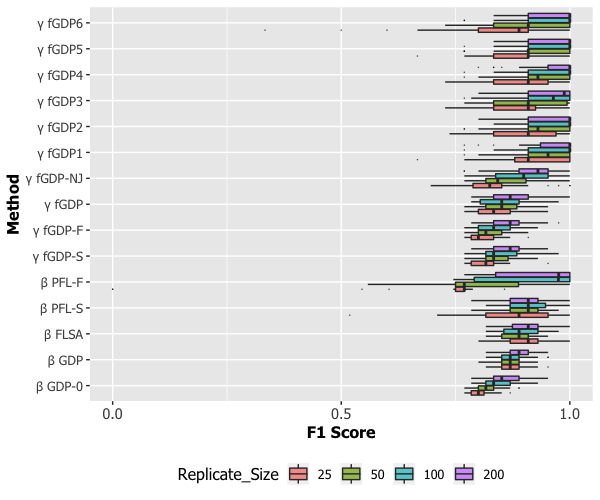}& \hspace{-0.3cm}\includegraphics[height=0.33\textwidth, width=0.33\textwidth]{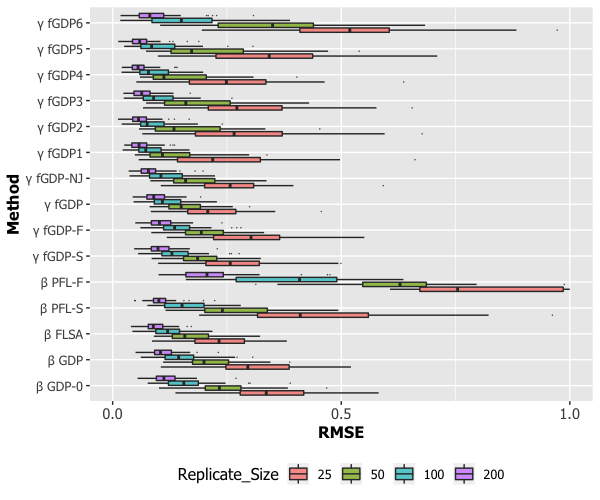}\\[-0.6cm]
\hspace{-0.3cm}\includegraphics[height=0.33\textwidth, width=0.33\textwidth]{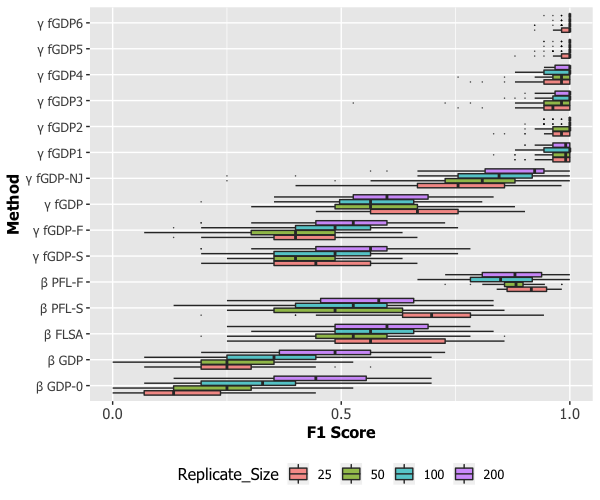} & \hspace{-0.3cm}\includegraphics[height=0.33\textwidth, width=0.33\textwidth]{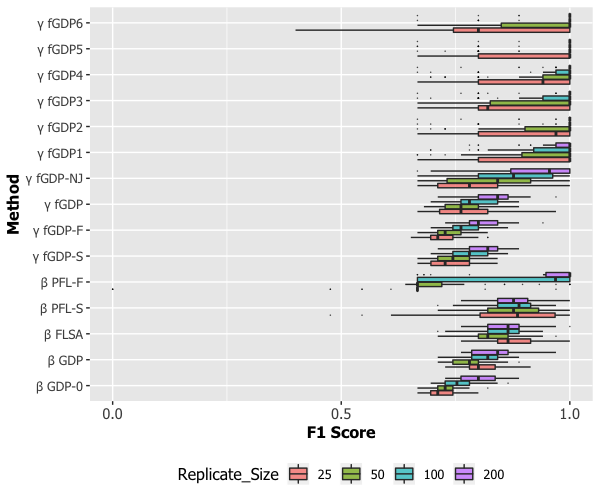}& \hspace{-0.3cm}\includegraphics[height=0.33\textwidth, width=0.33\textwidth]{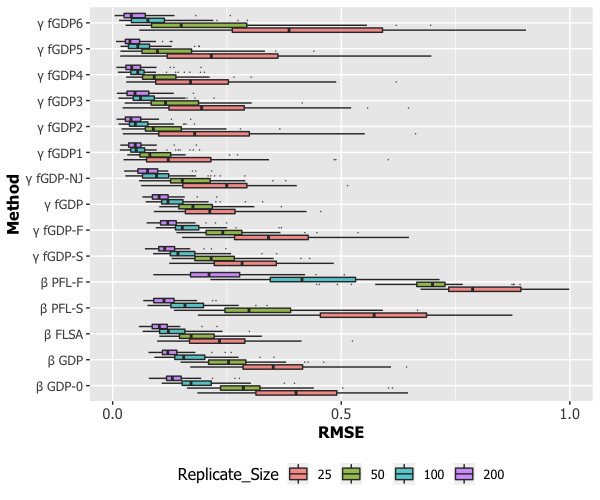}\\[-0.6cm]
\hspace{-0.3cm}\includegraphics[height=0.33\textwidth, width=0.33\textwidth]{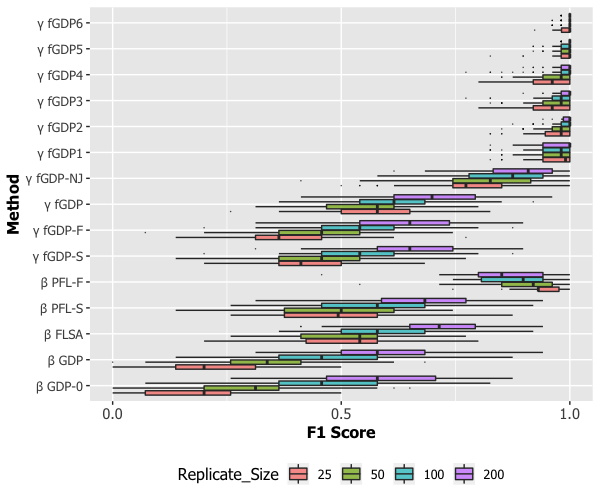} & \hspace{-0.3cm}\includegraphics[height=0.33\textwidth, width=0.33\textwidth]{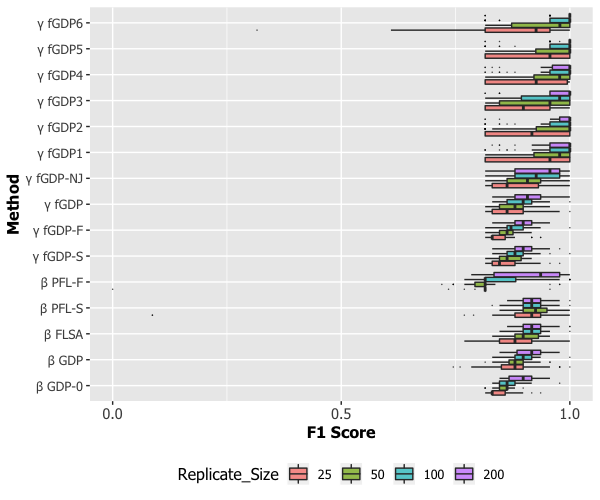}& \hspace{-0.3cm}\includegraphics[height=0.33\textwidth, width=0.33\textwidth]{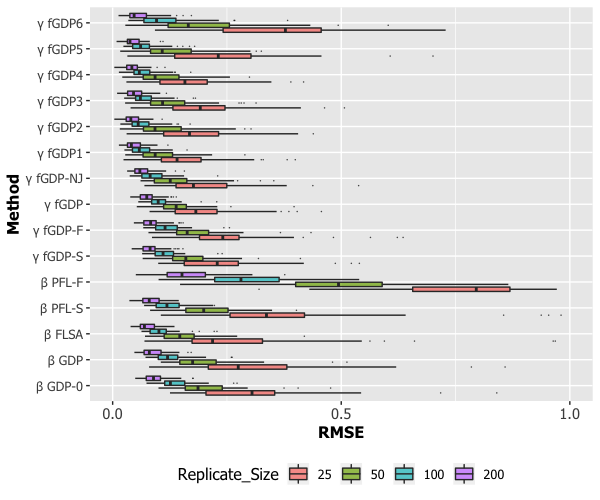}\\
\hspace{-0.1cm} \textrm{F1 score (fusion)} & \hspace{-0.1cm} \textrm{F1 score (selection)} & \hspace{-0.1cm} \textrm{RMSE} \\
\end{array}$}
\end{center}
 \vskip -0.10in
\caption{Boxplots of the F1 Scores and RMSE for $50$ simulations. Configurations (a)$\sim$(d) are shown on row $1\sim4$, respectively. The initial greek letter ($\beta$ or $\gamma$) indicates the parameter space of the baseline methods.}
\label{fig5: simulation}
\end{figure}

\subsubsection{Performance Evaluation}

We perform $50$ replications for every simulation scenario. To evaluate the performance, we use the F1 score = 2$\times$(precision$\times$ recall)/(precision+recall) as a metric.  In particular, for {selection}, we examine whether the non-zero regression coefficients in the $m$ leaf groups are detected. For {fusion}, we make $2^{h}-1$ binary decisions on whether the regression coefficients within the descendant set $\mathcal{D}_{s', \ell}$ are all equal, for every internal node $(s', \ell)\in\mathcal{I}_{h-1}$. We also provide the relative mean square error (RMSE) as a reference metric for signal recovery $\textrm{RMSE} = ||\hat{\bd{\beta}}-\bd{\beta}^{*}||_{\textrm{F}}/||\bd{\beta}^{*}||_{\textrm{F}}$. 

The results are illustrated in Figure \ref{fig5: simulation}. The results across rows demonstrate the adaptation ability of our PX-based fGDP approaches in recovering signals with multi-level and multiplex smoothness.  In general, the performance improves with the replicate size. When replicate size is relatively small, regularization helps boost the performance. Methods encouraging fusion produce better results in terms of F1 score (fusion) than those only encouraging selection. Note that by encouraging the expanded parameters on the root-to-leaf path to be sparse, the fGDP-S method is also able to encourage fusion in an indirect way.  

\subsubsection{Hyperparameter Sensitivity}

We observe that the default choice $\alpha = \eta = 1$ and the Normal-Jeffrey's choice $\alpha=\eta=0$ yield sub-optimal results. As the parameters in the PX space are redundant, a high level of shrinkage is required, demanding a small $\eta$ parameter. The PX-based fGDP approaches (fGDP1 $\sim$ fGDP6) with $\eta \in (10^{-3}, 10^{-1})$ compare favorably to the other methods in terms of both F1 scores (fusion and selection) and RMSE. Due to the adaptive shrinkage mechanism, the $\alpha$ parameters are less sensitive, we observed that $\alpha \in (0.5, 5)$ works reasonably well. In the original parameter space, the only method standing out in terms of the F1 scores is the fusion-dominated pairwise fused lasso (PFL-F) approach; however, this approach suffers from the inaccurate prior knowledge discussed in Section \ref{sec: operators}. As a result, it yields the worst RMSE performance among all the baselines considered. 
\begin{figure}[hbpt] 
\vskip -0.00in
\begin{center}
\scriptsize{
$\begin{array}{cccc}
\hspace{-0.31cm}\includegraphics[height=0.30\textwidth, width=0.25\textwidth]{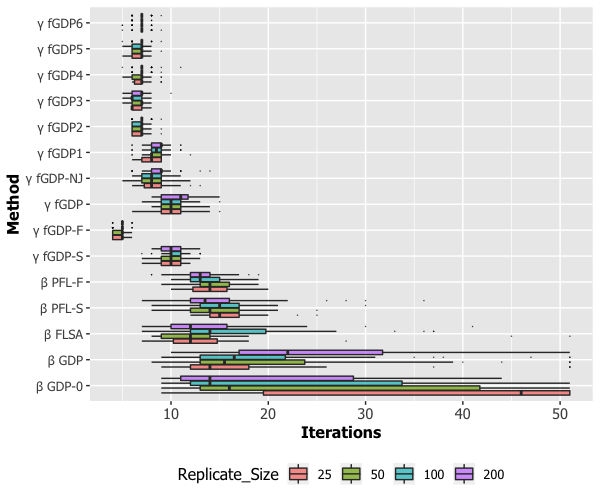} & \hspace{-0.31cm}\includegraphics[height=0.30\textwidth, width=0.25\textwidth]{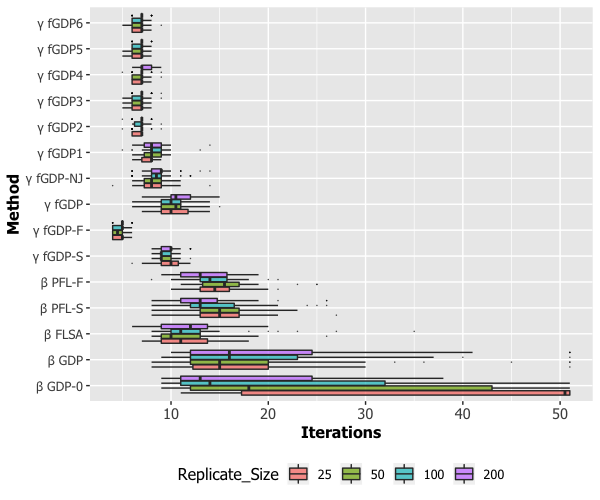}&
\hspace{-0.31cm}\includegraphics[height=0.30\textwidth, width=0.25\textwidth]{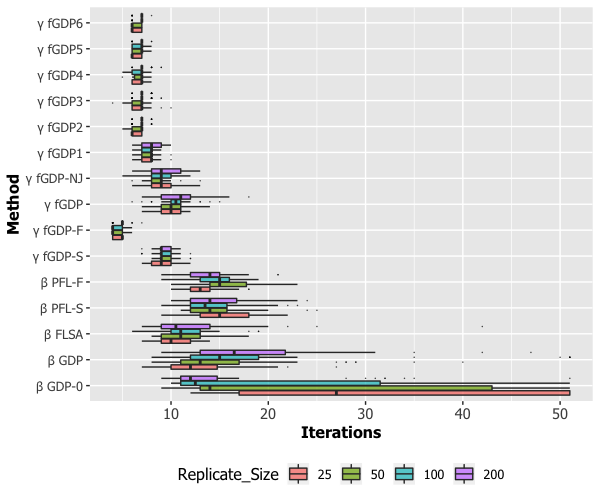} & \hspace{-0.31cm}\includegraphics[height=0.30\textwidth, width=0.25\textwidth]{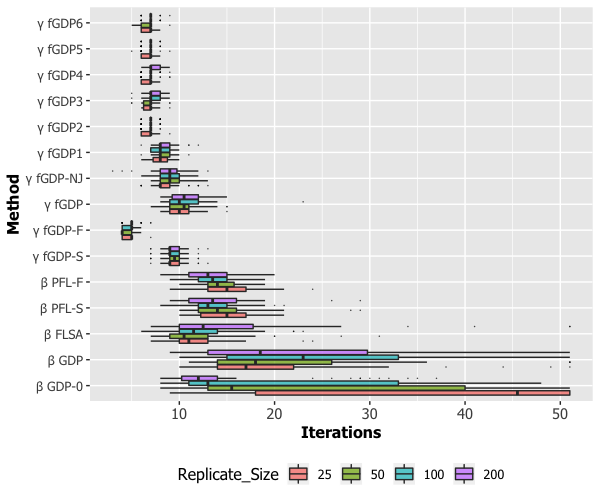}\\
\hspace{-0.1cm}  \textrm{Configuration (a)} & \hspace{-0.1cm}  \textrm{Configuration (b)} &\hspace{-0.1cm}  \textrm{Configuration (c)} & \hspace{-0.1cm}   \textrm{Configuration (d)}  \\
\end{array}$}
\end{center}
 \vskip -0.00in
\caption{The number of iterations needed for convergence with algorithms operating in the original parameter space (indicated by the initial greek letter $\beta$) or the expanded parameter space (indicated by the initial greek letter $\gamma$). }
\label{fig6: convergence}
\end{figure}

\subsubsection{Convergence}

We choose the convergence criteria to be $||\bd{\beta}^{(t)}-\bd{\beta}^{(t-1)}||_{\textrm{F}}<10^{-6}$, and for the PX-based method, this is $||\bd{D}\bd{\gamma}^{(t)}-\bd{D}\bd{\gamma}^{(t-1)}||_{\textrm{F}}<10^{-6}$. The quasi-Newton updates are performed with L-BFGS, with $1,000$ the maximum number of iterations allowed. We use the MATLAB solver  \emph{minFunc} \citep{schmidt2005minfunc}. Setting the maximum iterations of variational EM  to be $50$, the number of iterations until convergence is reported in Figure \ref{fig6: convergence}. As expected, the PX-based approaches converge much faster than the non-PX based approaches.

\begin{figure}[hbpt] 
\vskip -0.00in
\begin{center}
\footnotesize{
$\begin{array}{cc}
\hspace{-0.3cm}\includegraphics[height=0.26\textwidth, width=0.7\textwidth]{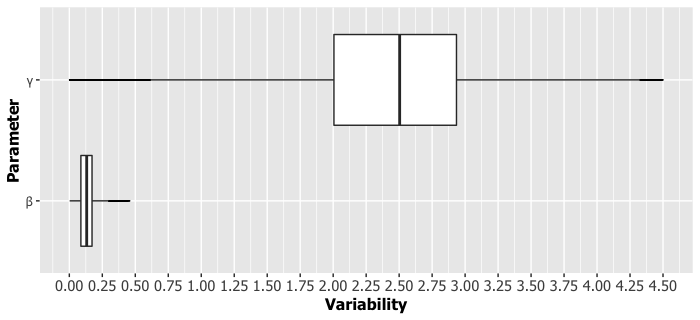} & \hspace{-0.3cm}\includegraphics[height=0.26\textwidth, width=0.3\textwidth]{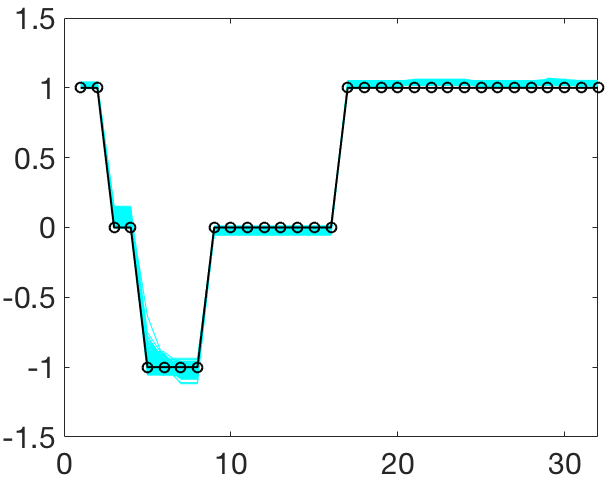} \\
\end{array}$}
\end{center}
 \vskip -0.00in
\caption{Simulation with different random initializations. Left panel: boxplots of pairwise distance between the estimated $\bd{\beta}$ or $\bd{\gamma}$ parameters across the $500$ runs. Right panel: estimations of $\bd{\beta}$ (cyan lines) comparing against the ground truth (black circled line).}
\label{fig7: random_ini}
\end{figure}

The PX-based approaches only converge to one of the many local optima. We repeated the fGDP2 approach with hyper-parameters $\alpha_{1} = \alpha_{2} = 1$, $\eta_{1} = \eta_{2} = 0.01$ with $500$ different random initializations. For every pair of different runs $a, b\in \{1, \hdots, 500\}$ and $a\neq b$, we calculate the pairwise distances $||\hat{\bd{\beta}}_{a}-\hat{\bd{\beta}}_{b}||_{\textrm{F}}$ and $||\hat{\bd{\gamma}}_{a}-\hat{\bd{\gamma}}_{b}||_{\textrm{F}}$ as a metric for variability. The results under configuration (d) with $n = 200$ are summarized in Figure \ref{fig7: random_ini}, which show that the algorithm converges to a large number of different local optimal solutions in the auxiliary space; however, when reducing to the original space, they all map to solutions with comparable quality lying within a close neighborhood around the ground truth $\bd{\beta}^{*}$.

\subsection{Supervised Dimension Reduction of Soccer Passing Networks}\label{sec6b}
As a team sport, soccer is characterized by its free-flowing nature \citep{gudmundsson2017spatio}. The spatial interaction networks provide a concise abstraction of the team play on the pitch, which capture the essence of soccer as an invasion-territorial sport. Our spinlets approach provides a useful tool for further reducing complexity, taking advantage of both the passing network structure and the response relevance. On the FIFA World Cup 2018 dataset, we {\color{black}first consider two types of response} in the two tasks below: 
\be
\item  {\bf Task 1}: Team performance measured by the goal difference (i.e., goal scored minus goal conceded)  in the $128$ game plays (unique team-game pairs).
\item  {\bf Task 2}: The urgency and tiredness status of the game,  indicated by whether the data are collected after $70$ minutes. This division of game play into two game phases results in $128\times 2 = 256$ replicates with binary responses. 
\ee

\begin{figure}[hbpt] 
\vskip -0.00in
\begin{center}
\footnotesize{
$\begin{array}{ccc}
\hspace{-0.3cm}\includegraphics[height=0.26\textwidth, width=0.33\textwidth]{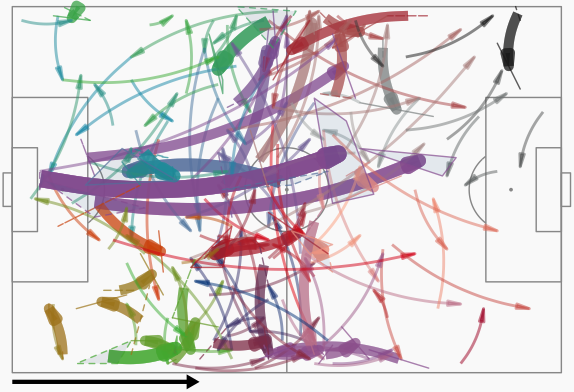} & \hspace{-0.3cm}\includegraphics[height=0.26\textwidth, width=0.33\textwidth]{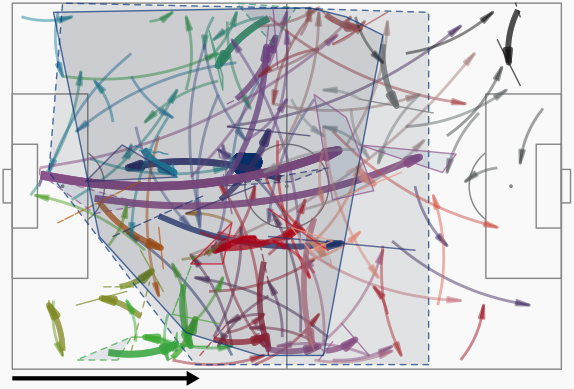}& \hspace{-0.3cm}\includegraphics[height=0.26\textwidth, width=0.33\textwidth]{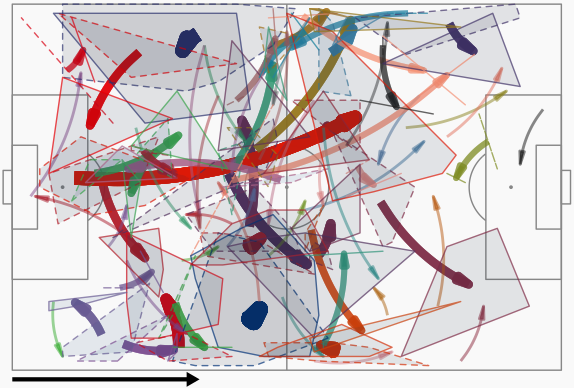}\\[0.2em]
\hspace{-0.3cm}\includegraphics[height=0.26\textwidth, width=0.33\textwidth]{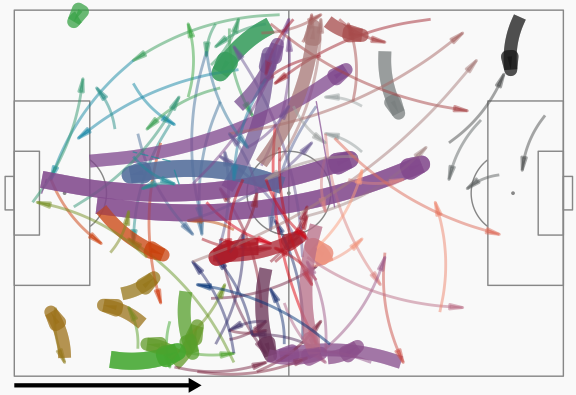} & \hspace{-0.3cm}\includegraphics[height=0.26\textwidth, width=0.33\textwidth]{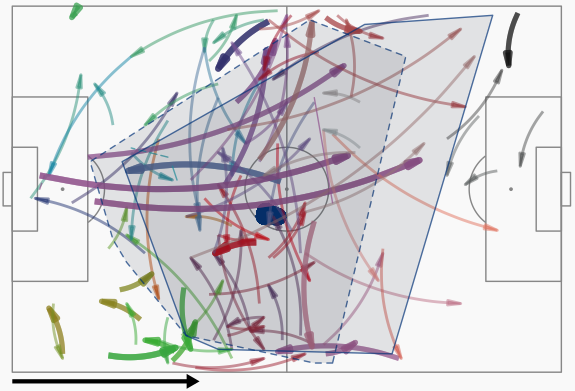}& \hspace{-0.3cm}\includegraphics[height=0.26\textwidth, width=0.33\textwidth]{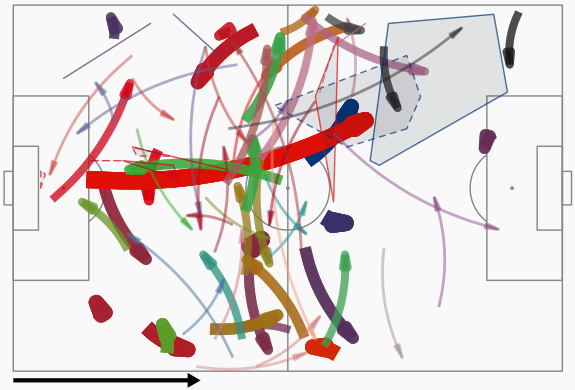}\\[0.2em]
\hspace{-0.3cm}\includegraphics[height=0.26\textwidth, width=0.33\textwidth]{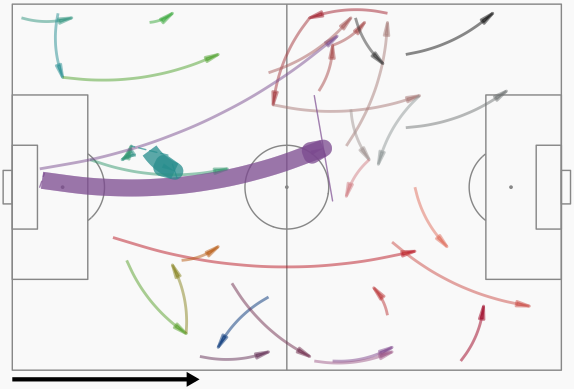} & \hspace{-0.3cm}\includegraphics[height=0.26\textwidth, width=0.33\textwidth]{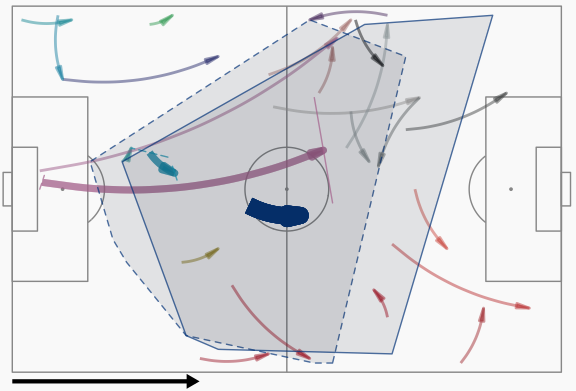}& \hspace{-0.3cm}\includegraphics[height=0.26\textwidth, width=0.33\textwidth]{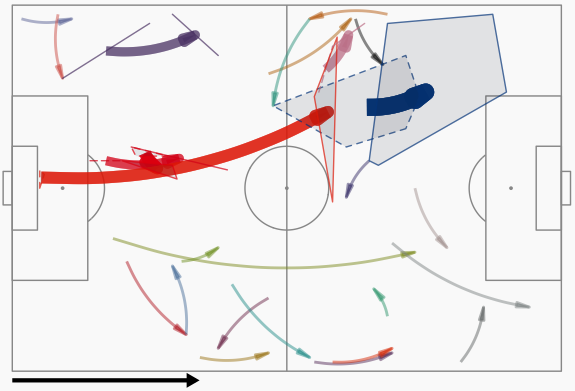}\\[0.2em]
\hspace{-0.1cm} \textrm{(a) rMETIS} & \hspace{-0.1cm} \textrm{(b) S3O} & \hspace{-0.1cm} \textrm{(c)  Spinlets (our method)}  \\[0.2em]
\end{array}$}
\end{center}
 \vskip -0.20in
\caption{SPIN representations of France in the World Cup Final 2018, under the supervision of goal difference (first row), or under the supervision of game phase indicator in the first $70$ minutes (second row), and after $70$ minutes (third row).  The colored arrows represent the grouped POs with width proportional to the count of occurrences  in the group. For the sake of visualization, the origins (dashed) and destinations (solid) of the grouped POs are indicated by the convex hulls (shaded region) with the grouped POs located at the centroids of the polygons. }
\label{fig9a: viz}
\end{figure}

\begin{figure}[hbpt] 
\vskip -0.00in
\begin{center}
\footnotesize{
$\begin{array}{ccc}
\hspace{-0.3cm}\includegraphics[height=0.26\textwidth, width=0.33\textwidth]{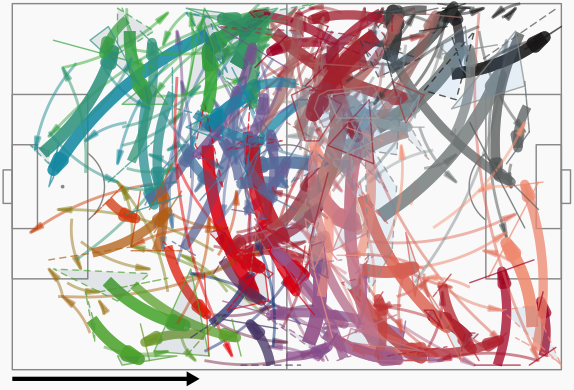} & \hspace{-0.3cm}\includegraphics[height=0.26\textwidth, width=0.33\textwidth]{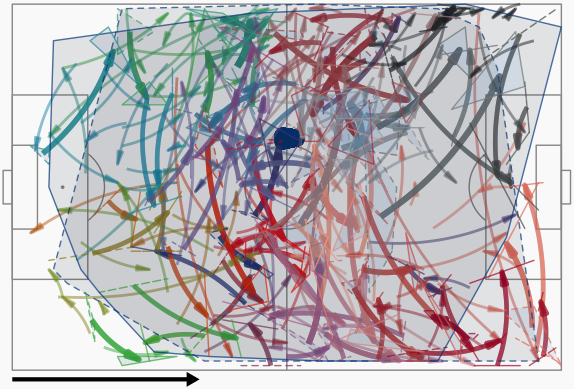}& \hspace{-0.3cm}\includegraphics[height=0.26\textwidth, width=0.33\textwidth]{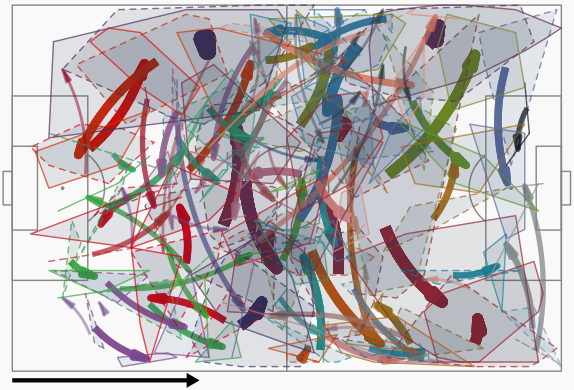}\\[0.2em]
\hspace{-0.3cm}\includegraphics[height=0.26\textwidth, width=0.33\textwidth]{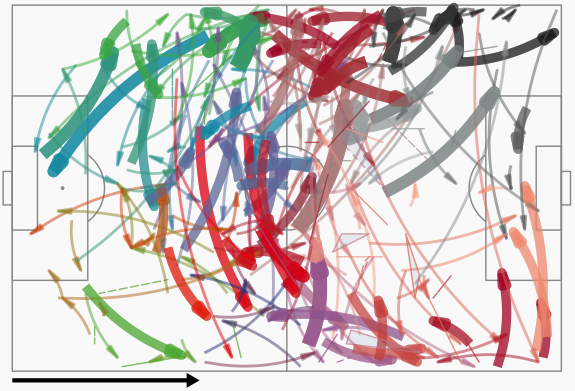} & \hspace{-0.3cm}\includegraphics[height=0.26\textwidth, width=0.33\textwidth]{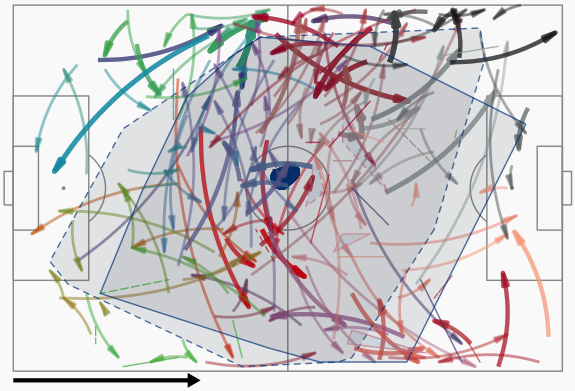}& \hspace{-0.3cm}\includegraphics[height=0.26\textwidth, width=0.33\textwidth]{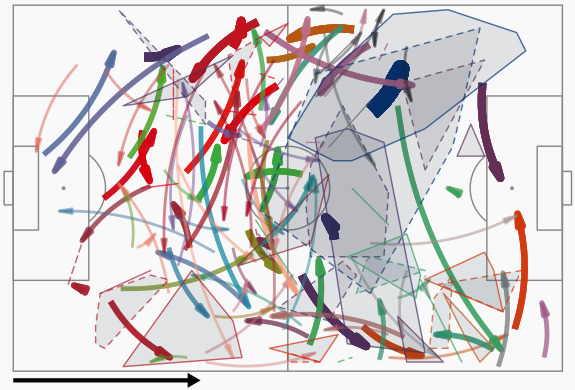}\\[0.2em]
\hspace{-0.3cm}\includegraphics[height=0.26\textwidth, width=0.33\textwidth]{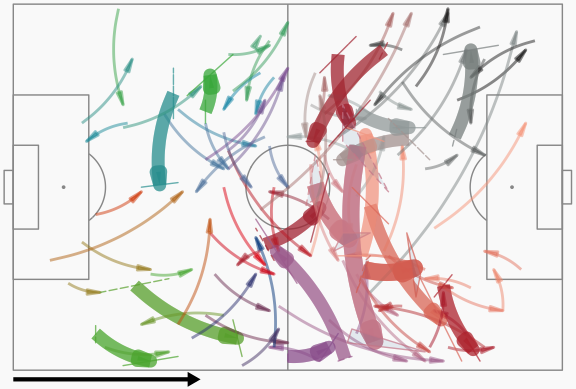} & \hspace{-0.3cm}\includegraphics[height=0.26\textwidth, width=0.33\textwidth]{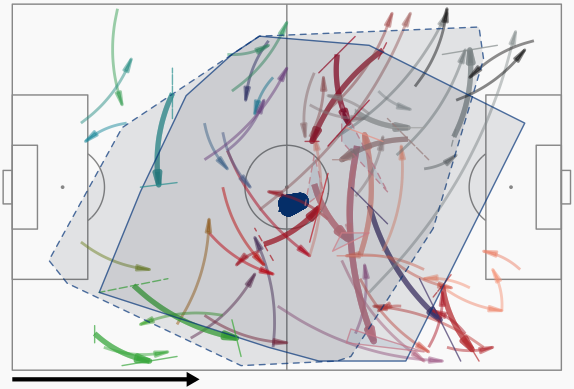}& \hspace{-0.3cm}\includegraphics[height=0.26\textwidth, width=0.33\textwidth]{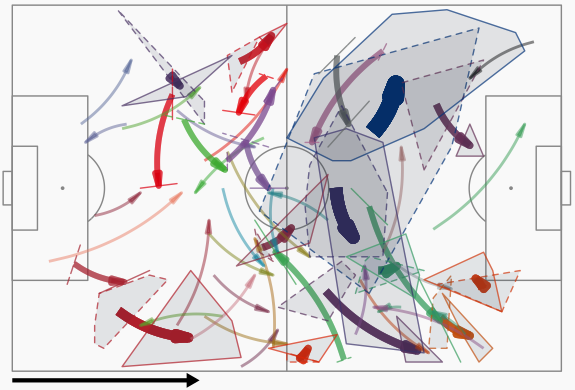}\\[0.2em]
\hspace{-0.1cm} \textrm{(a) rMETIS} & \hspace{-0.1cm} \textrm{(b) S3O} & \hspace{-0.1cm} \textrm{(c)  Spinlets (our method)}  \\[0.2em]
\end{array}$}
\end{center}
 \vskip -0.10in
\caption{SPIN representations of Croatia in the World Cup Final 2018, under the supervision of goal difference (first row), or under the supervision of game phase indicator in the first $70$ minutes (second row), and after $70$ minutes (third row).  The colored arrows represent the grouped POs with width proportional to the count of occurrences  in the group. For the sake of visualization, the origins (dashed) and destinations (solid) of the grouped POs are indicated by the convex hulls (shaded region) with the grouped POs located at the centroids of the polygons.}
\label{fig9b: viz}
\end{figure}

We compare three algorithms: (1) {\bf rMETIS} with $h=9$, (2) a single-scale selection-only  ({\bf S3O}) approach with the default GDP prior on $\bd{\beta}$, and (3) {\bf Spinlets} with the fGDP2 prior on $\bd{\gamma}$. The threshold for the regression coefficients to be considered as (approximately) equal is set to be $0.005$. The {\bf rMETIS} algorithm provides a common starting point of CG representation for both {\bf S3O} and {\bf Spinlets}.   Interestingly, the same game can be seen with different eyes, when the practitioner picks a different response variable. In task 1 we consider team performance measured by the goal differences, while in task 2 we consider whether the passing network is collected in a relatively late phase of the game, when the time is running out and the players are getting physically and mentally tired. Figure \ref{fig9a: viz} and \ref{fig9b: viz} illustrate the reductive representations of exactly the same games with two different types of response. {\color{black}Spinlets merges functionally similar passes together in the same group, removes the ones irrelevant to the response, and produces visual outputs that are readily interpretable to the human eye.}

\begin{figure}[hbpt] 
\vskip -0.00in
\begin{center}
\footnotesize{
$\begin{array}{cc}
\hspace{-0.3cm}\includegraphics[height=0.34\textwidth, width=0.49\textwidth]{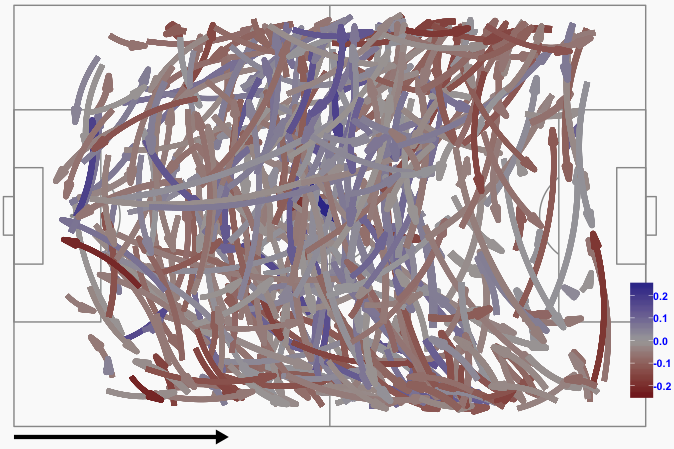} &
\hspace{-0.3cm}\includegraphics[height=0.34\textwidth, width=0.49\textwidth]{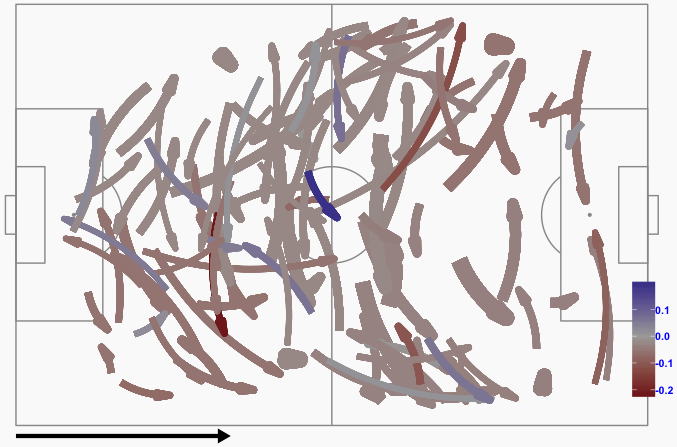}\\[0.2em]
\hspace{-0.3cm} \textrm{(a) Task 1 (goal difference): S3O} & \hspace{-0.3cm} \textrm{(b) Task 1 (goal difference): Spinlets}\\[0.2em]
\hspace{-0.3cm}\includegraphics[height=0.34\textwidth, width=0.49\textwidth]{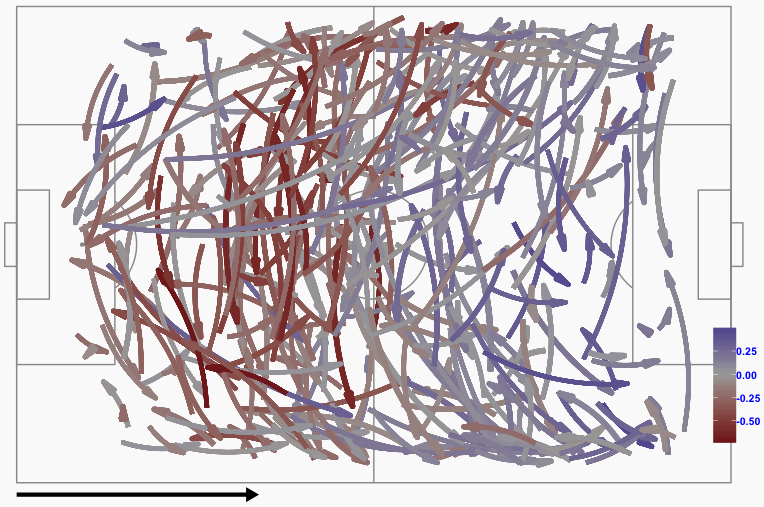} &
\hspace{-0.3cm}\includegraphics[height=0.34\textwidth, width=0.49\textwidth]{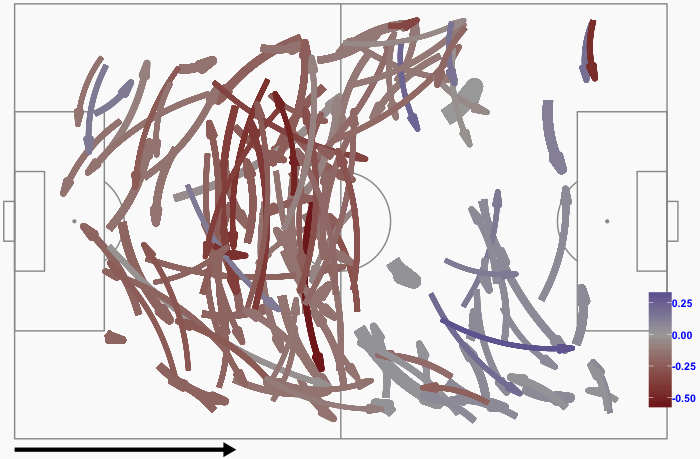}\\[0.2em]
\hspace{-0.3cm} \textrm{(c) Task 2 (game phase): S3O} & \hspace{-0.3cm} \textrm{(d) Task 2 (game phase): Spinlets}\\[0.2em]
\end{array}$}
\end{center}
 \vskip -0.10in
\caption{The estimated regression coefficients $\hat{\bd{\beta}}$ in the two tasks.}
\label{fig12a: beta}
\end{figure}

The magnitude of the estimated regression coefficients $\hat{\bd{\beta}}$ indicates the strength of association with the response. {\color{black}For the sake of clarity, we display them in separate plots in Figure \ref{fig12a: beta}.} With both selection and fusion,  {\bf Spinlets} produces more concise representations than {\bf S3O} and allows for non-uniform resolutions. {\color{black}Note that the number of passes in the different partition sets defines a passing strategy --- this strategy can be related to the competition outcome or a situational game factor. These plots find the common pattern from all the games.} For example, Figure \ref{fig12a: beta} (b) suggests that the cross-passes from the two wings seem relatively inefficient in winning the game, or adopted more by the losing team. In contrast, Figure \ref{fig12a: beta} (d) indicates that teams tend to control the midfield more in the first $70$ minutes,  while pushing for the goal by passing more in the opposing half after $70$ minutes.

\begin{figure}[hbpt] 
\vskip -0.00in
\begin{center}
\footnotesize{
$\begin{array}{c}
\hspace{-0.3cm}\includegraphics[height=0.3\textwidth, width=0.85\textwidth]{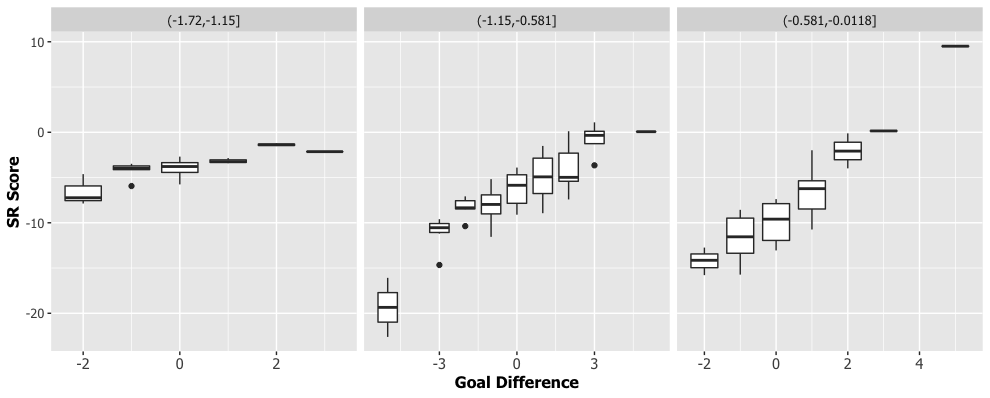}\\
\hspace{-0.3cm} \textrm{(a) Task 1 (goal difference): S3O} \\
\hspace{-0.3cm}\includegraphics[height=0.3\textwidth, width=0.85\textwidth]{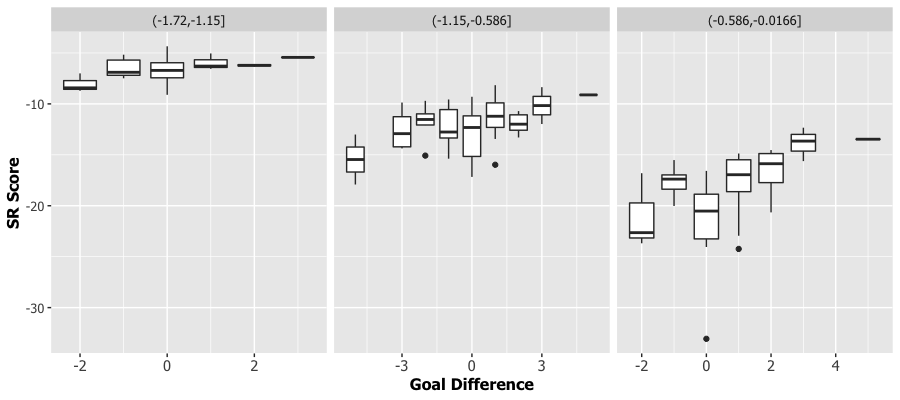}\\
\hspace{-0.3cm} \textrm{(b) Task 1 (goal difference): Spinlets} \\
\hspace{-0.3cm}\includegraphics[height=0.3\textwidth, width=0.85\textwidth]{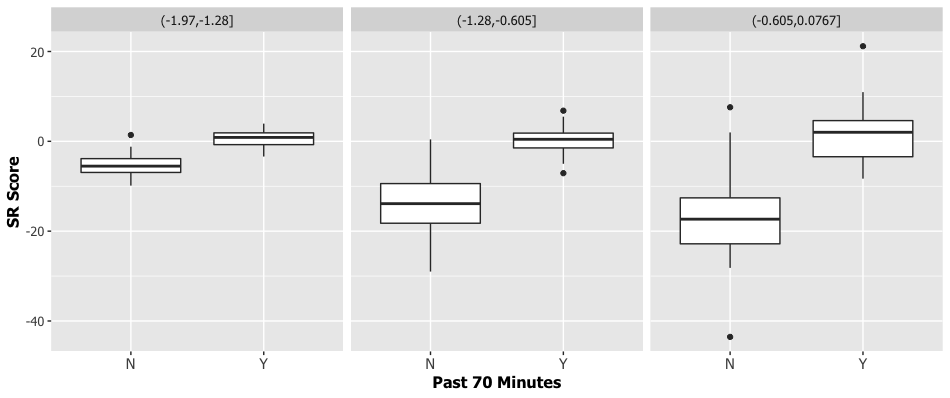}\\
\hspace{-0.3cm} \textrm{(c) Task 2 (game phase): S3O} \\
\hspace{-0.3cm}\includegraphics[height=0.3\textwidth, width=0.85\textwidth]{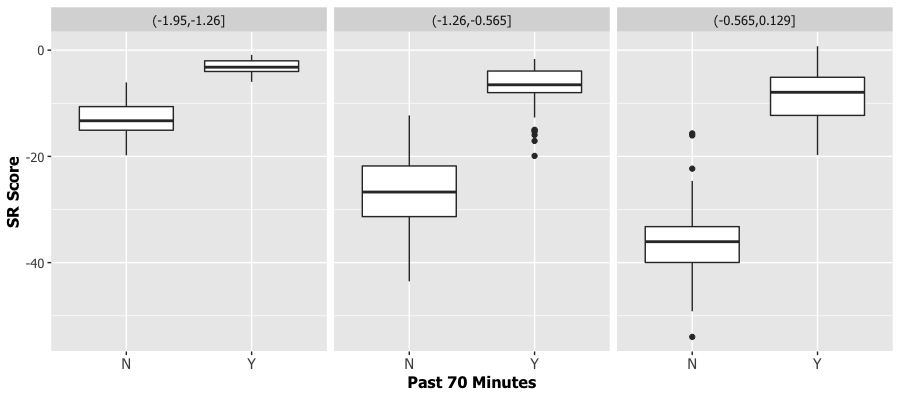}\\
\hspace{-0.3cm} \textrm{(d) Task 2 (game phase): Spinlets} \\
\end{array}$}
\end{center}
 \vskip -0.10in
\caption{SDR score v.s. response in the two tasks, separated into $3$ groups according to the estimated mean of the replicate-specific random effects.}
\label{fig13: sdr-score}
\end{figure}

To illustrate the {\color{black}information preserved in} the reductive representation, we plot the SDR score {\color{black}$R_{\hat{\bd{\beta}}}(\bd{x}_{i})$} versus {\color{black}$y_{i}$} in Figure \ref{fig13: sdr-score}.  For the PIR mixed model, the sufficiency of the SDR score holds conditionally on the replicate-specific random effects $[a, a+\bd{b}]$, so we cut the estimated mean of random effects $[{\zeta}_{a}, {\zeta}_{a}+\bd{\zeta}_{b}]$ into three intervals, and present the results in three separate panels. Both  {\bf S3O} and {\bf Spinlets} are regularized approaches, in which fitting the data is not the only goal. Comparing to {\bf S3O}, {\bf Spinlets} produces more parsimonious results with little or no sacrifice in discriminative power, as shown in Figure \ref{fig13: sdr-score}.

\newpage
\subsection{{\color{black}Predictive Powers of the Vectorial Representation}}\label{sec6c}
{\color{black}Different vectorial representation of the SPIN data can be obtained via different operations on the partitioning tree. For example, the {\bf rMETIS} method cuts the tree at a single height, the {\bf S3O} method prunes irrelevant leaf groups, and our {\bf Spinlets} method enables both deletion and fusion across multiple scales. Although not the main focus of this article, the predictive powers of the vectorial representations induced by these methods are also evaluated. In the dimension reduction step, we first perform {\bf rMETIS} with $h=9$ for all the data, and then split them into a training set and a testing set in a ratio of $90:10$. To refine the initial vectorial representation, the tree is cut at $s = 7$ or $s=8$ unsupervisedly,  or trimmed by {\bf S3O} or {\bf Spinlets} with the number of goals scored as supervision. The vectorial representations on the testing set can be determined accordingly. 

Next, these vectorial representations are plugged in a forward Poisson regression model as predictors with the number of goals scored as the response. Specifically, we use the {\bf glmnet} implementation \citep{glmnet} of the  Poisson log-linear regression model penalized by lasso ($\alpha = 1$), elastic-net ($\alpha = 0.5$), and ridge ($\alpha = 0$). For each case, we perform $10$-fold cross-validation on the training data, and evaluate the performance on the held-out test data, both using the mean absolute error (MAE) metric. We conduct $100$ replicated simulations and summarize the mean and standard deviation of MAE  in Table \ref{Table: MAE}. Predicting the number of goals scored from passing strategies is, admittedly, a challenging task, and our spinlets method gives slightly improved generalization performance.   

\begin{table}[bpht]
\begin{center}
\caption{Mean and standard deviation (subscripts) of MAEs.}\label{Table: MAE}
\begin{tabular}{|l|l|l|l|l|l|l|}
\hline
               & \multicolumn{2}{c|}{Lasso $(\alpha = 1)$} & \multicolumn{2}{c|}{Elastic-net $(\alpha = 0.5)$} & \multicolumn{2}{c|}{Ridge $(\alpha = 0)$} \\ \hline
Method         & \multicolumn{1}{c|}{Training}       & \multicolumn{1}{c|}{Testing}       &  \multicolumn{1}{c|}{Training}        & \multicolumn{1}{c|}{Testing}         & \multicolumn{1}{c|}{Training}       & \multicolumn{1}{c|}{Testing}        \\ \hline
rMETIS (h=7)   & $0.851_{0.046}$          & $0.908_{0.183}$       & $0.845_{0.058}$           & $0.910_{0.184}$          & $0.756_{0.063}$          & $0.899_{0.182}$         \\ \hline
rMETIS (h=8)   & $0.861_{0.068}$          & $0.924_{0.183}$        & $0.857_{0.073}$           & $0.934_{0.192}$          & $0.848_{0.085}$          & $0.921_{0.177}$         \\ \hline
rMETIS (h=9)   & $0.712_{0.112}$          & $0.913_{0.189}$         & $0.702_{0.120}$           & $0.909_{0.188}$          & $0.580_{0.114}$          & $0.901_{0.180}$         \\ \hline
S3O       & $0.699_{0.128}$          & $0.919_{0.208}$         & $0.694_{0.116}$           & $0.910_{0.190}$          & $0.476_{0.031}$          & $0.879_{0.190}$         \\ \hline
Spinlets  & $0.574_{0.078}$          & $0.909_{0.208}$         & $0.564_{0.071}$           & $0.899_{0.205}$          & $0.611_{0.057}$          & $0.867_{0.198}$         \\ \hline
\end{tabular}
\end{center}
\end{table}
}

\section{Discussion}\label{sec7}
In this article, we have introduced spinlets---a supervised dimension reduction method for spatial interaction networks using information on the spatial locations of each pass and also a response variable in constructing a multiscale representation.  Particularly, the dimension reduction is conducted via a top-down partitioning of the similarity graph and a bottom-up pruning of the partition tree. Instead of cutting at a single height of a given tree, we select multiple tree heights for different branches of the tree adaptively, which yields representations with mixed granularities. The regularization prevents the information preserved in the lower-dimensional representation from being dominated by the supervisory signal without enough conformity to the spatial network data structures. In addition, our approach can be interpreted as an empirical Bayes approach, which estimates a hierarchical tree organization of the data in a first stage. 

Besides the sports application studied in this article, our spinlets approach can accommodate massive-scale network predictors or temporally-indexed predictors with high sampling rate; both are pressing needs in neuroscience. Potentially further improvements on the flexibility and utility of spinlets are possible, such as zero-inflated variants, SDR with multiple responses and covariate adjustment.

 \section{Acknowledgment}
This research was supported by grant W911NF-16-1-0544 from the U.S. Army Research Institute for the Behavioral and Social Sciences (ARI).
The authors thank StatsBomb for making the data freely available in public for academic research.  The authors would also like to thank Li Ma and Andr\'{e}s Felipe Barrientos for inspirational discussions.

\appendix

 \section{Preprocessing---Recursive METIS (rMETIS)}\label{appendix1}

\begin{figure}[hbpt] 
\vskip -0.00in
\begin{center}
\footnotesize{
$\begin{array}{c}
\hspace{-0.3cm}\includegraphics[height=0.42\textwidth, width=0.95\textwidth]{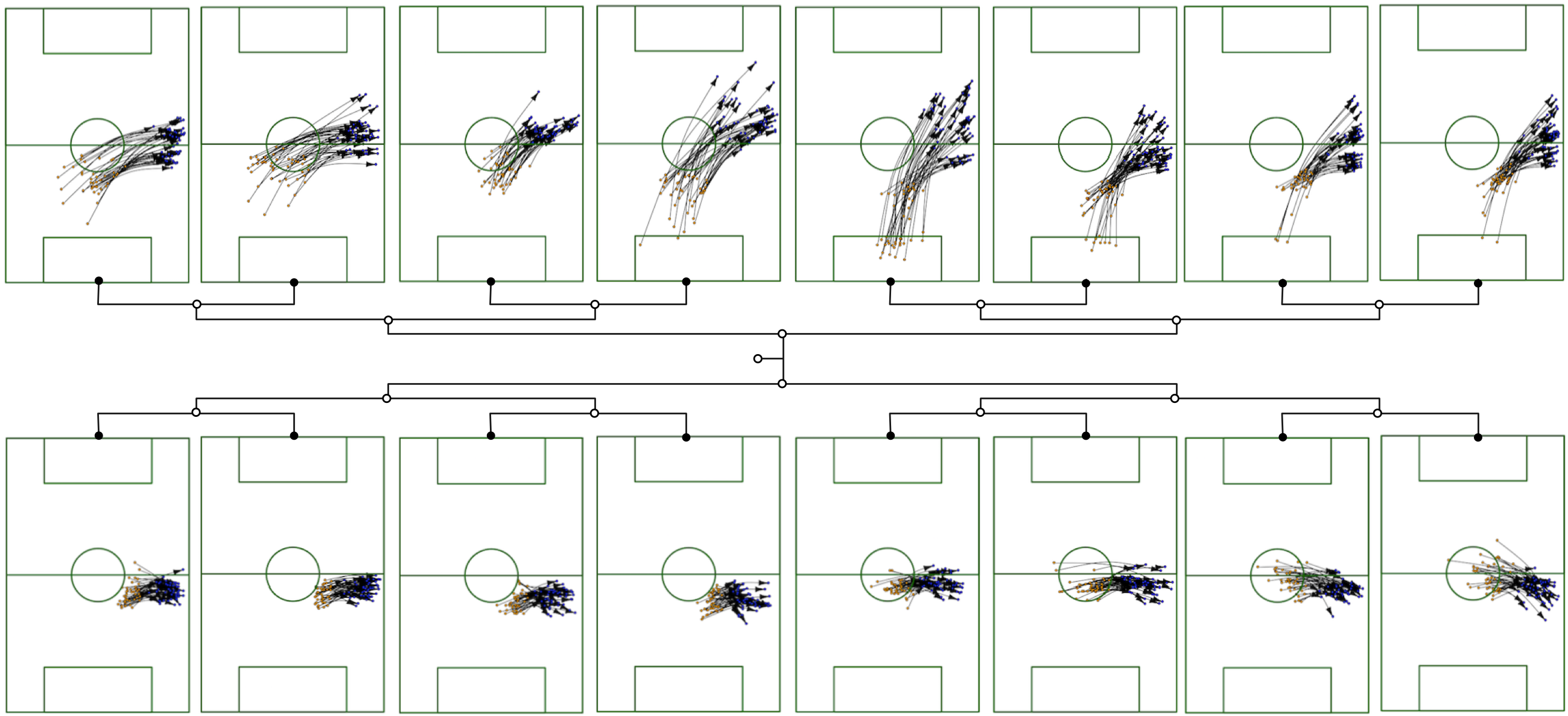}\\
\end{array}$}
\end{center}
 \vskip -0.10in
\caption{A sub-branch of the partition tree with $16$ leave nodes (indicated by solid dots) produced by the rMETIS algorithm. Every internal node (indicated by hollow dots) on the coarse scale is connected with two child nodes on the finer scale.}
\label{fig4: tree16}
\end{figure}

To concisely represent the proximity information of POs (e.g., soccer passes), we define a tree structure in three steps: $(i)$ choose the Euclidean distance metric between pairs of POs and compute a $Q\times Q$ distance matrix, $Q = \sum_{i}q_{i}$; $(ii)$ construct a sparse similarity graph of POs $\mathcal{G}$ by considering $K$ nearest neighbors; $(iii)$ build a partition tree $\mathcal{T}_{h}$ via recursively applying the METIS partitioning algorithm \citep{karypis1998fast} on $\mathcal{G}$. We construct a full binary tree via recursive METIS (rMETIS). In each step, a set of POs is split into two disjoint subsets. Figure \ref{fig4: tree16} illustrates a sub-branch of $\mathcal{T}_{10}$ with all assigned POs in the $16$ groups plotted. 

The $Q\times Q$ distance matrix is constructed approximately by the k-d tree nearest neighbor search, which has $O(Q)$ worst case complexity. Second, the rMETIS algorithm assigns the POs into $m$ primary groups. After $h$ phases, the set of POs $\mathcal{P}$ is partitioned into $m=2^{h}$ leaf groups on $\mathcal{T}_{h}$. The complexity of the bisection algorithm is ${O}(|E|\log{m})$, where $|E|$ is the number of edges in the similarity graph. For $Q=49,988$, constructing the nearest neighbor set (with $K=1,500$ nearest neighbors) requires $15.97$ seconds,  computing the similarity matrix costs $68.58$ seconds, and running the rMETIS algorithm (with tree depth $h=9$) takes $52.87$ seconds. 

\section{Gradients in the M-Step}\label{appendix: a}

The variational parameters $\bd{\zeta}$ and $\bd{\kappa}$ should be chosen to make $\underline{\ell}(\bd{\gamma}, \bd{\omega}, \bd{\zeta}, \bd{\kappa})$ as close as possible to $\ell(\bd{\gamma}, \bd{\omega})$. Given $q(\bd{\Psi})$ and denoting $\mathcal{F}(q(\bd{\Psi}), \bd{\gamma}, \bd{\omega})  = Q(\bd{\gamma}, \bd{\omega}, \bd{\zeta}, \bd{\kappa})$, we have 
\bi
\item {\bf Gradients of $\bd{\gamma}$}:
Denoting the gradients of $Q(\bd{\gamma}, \bd{\omega}, \bd{\zeta}, \bd{\kappa})$ over $\gamma_{l}$ as $D_{\gamma_{l}}{Q}$, we have
\eq{
D_{\gamma_{l}}{Q} & = \sum_{i=1}^{n} \sum_{j=1}^{m}\epsilon_{i,j}y_{i}d_{j,l} - [\wt{\bd{\Lambda}}^{(t)}\bd{\gamma}]_{l}, \quad \epsilon_{i,j} = x_{i, j} - \tilde{x}_{i, j}, 
}
where $\tilde{x}_{i, j} = t_{i}\exp{[\zeta^{a}+\zeta^{b}_{i}+\zeta^{c}_{j}+\frac{1}{2}(\kappa^{a}+\kappa^{b}_{i}+\kappa^{c}_{j})+y_{i}\sum_{l=1}^{L}{d}_{j,l}{\gamma}_{l}]}$. 
\item {\bf Gradients of $\bd{\zeta}$}: Denoting the gradients of $Q(\bd{\gamma}, \bd{\omega}, \bd{\zeta}, \bd{\kappa})$ over $\zeta^{a}$, $\zeta_{i}^{b}$, and $\zeta_{j}^{c}$ as $D_{\zeta^{a}}{Q}$, $D_{\zeta_{i}^{b}}{Q}$, and $D_{\zeta_{j}^{c}}{Q}$ respectively, we have
\eq{
D_{\zeta^{a}}{Q}  & = -\frac{\zeta^{a}}{\omega_{a}}+\sum_{i=1}^{n}\sum_{j=1}^{m}\epsilon_{i,j}, \quad  D_{\zeta_{i}^{b}}{Q}   = -\frac{\zeta_{i}^{b}}{\omega_{b}}+\sum_{j=1}^{m}\epsilon_{i,j},  \quad  D_{\zeta_{j}^{c}}{Q}  = -\frac{\zeta_{j}^{c}}{\omega_{c}}+\sum_{i=1}^{n}\epsilon_{i,j}.
}
\item {\bf Gradients of $\bd{k}$}: To ensure positivity of the variance parameters $\bd{\kappa}$,  we apply the $\log$ reparameterization on $\bd{\kappa}$. Denoting the gradients of $Q(\bd{\gamma}, \bd{\omega}, \bd{\zeta}, \bd{\kappa})$ over $k^{a} = \log{(\kappa^{a})}$, $k_{i}^{b} = \log{(\kappa_{i}^{b})}$, and $k_{j}^{c}= \log{(\kappa_{j}^{c})}$ as $D_{k^{a}}{Q}$, $D_{k_{i}^{b}}{Q}$, and $D_{k_{j}^{c}}{Q}$ respectively, we have
\eq{
D_{k^{a}}{Q}  & = -\frac{\exp{(k^{a})}}{2\omega_{a}}+\frac{1}{2}-\frac{1}{2}\sum_{i=1}^{n}\sum_{j=1}^{m}\tilde{x}_{i, j}\exp{(k^{a})}, \cr  D_{k_{i}^{b}}{Q}  & = -\frac{\exp{(k_{i}^{b})}}{2\omega_{b}}+\frac{1}{2}-\frac{1}{2}\sum_{j=1}^{m}\tilde{x}_{i, j}\exp{(k_{i}^{b})},  \cr  D_{k_{j}^{c}}{Q}  &= -\frac{\exp{(k_{j}^{c})}}{2\omega_{c}}+\frac{1}{2}-\frac{1}{2}\sum_{i=1}^{n}\tilde{x}_{i, j}\exp{(k_{j}^{c})}. 
}
\ei

\section{GDP Family Priors  on $\bd{\beta}$}\label{appendix: b}
\paragraph{The GDP Prior}
The GDP penalty term is $\log{p(\bd{\beta})} = \sum_{j=1}^{m}\log{p({\beta}_{j})} = \sum_{j=1}^{m}[-\log{(2\xi)}-(\alpha+1)\log{(1+\frac{|{\beta}_{j}|}{\alpha \xi})]}$. 
In the E-Step, we have $\langle{\rho_{j}}\rangle  ={(\alpha+1)}/{[|{\beta_{j}}^{(t)}|({|{\beta_{j}}^{(t)}|}+\eta)]}$.  In the M-step, the gradient w.r.t $\beta_{j}$ can be written as,  
\eq{
D_{\beta_{j}}{Q} & = \sum_{i=1}^{n} \epsilon_{i,j}y_{i} - \langle{\rho_{j}}\rangle \beta_{j},\quad \epsilon_{i,j} = x_{i, j} - \tilde{x}_{i, j},
}
where $\tilde{x}_{i, j}  := t_{i}\exp{[\zeta^{a}+\zeta^{b}_{i}+\zeta^{c}_{j}+\frac{1}{2}(\kappa^{a}+\kappa^{b}_{i}+\kappa^{c}_{j})+y_{i}\beta_{j}]}$ and the gradients w.r.t other parameters do not change.

\paragraph{GDP-based FLSA}
The GDP-based FLSA penalty term is  
\eq{
\log{p(\bd{\beta})} & = \sum_{j=1}^{m}\log{p({\beta}_{j})} = \sum_{j=1}^{m}\bigg[-\log{(2\xi_{1})}-(\alpha_{1}+1)\log{\bigg(1+\frac{|{\beta}_{j}|}{\alpha_{1} \xi_{1}}}\bigg)\bigg]\cr 
& + \sum_{j'=1}^{m-1}\bigg[-\log{(2\xi_{2})}-(\alpha_{2}+1)\log{\bigg(1+\frac{|{\beta}_{j'}-{\beta}_{j'+1}|}{\alpha_{2} \xi_{2}}}\bigg)\bigg].
}
In the E-Step, we have 
\eq{\langle{\rho_{j}}\rangle  ={(\alpha_{1}+1)}/{[|{\beta_{j}}^{(t)}|({|{\beta_{j}}^{(t)}|}+\eta_{1})]}, \quad \langle{\upsilon_{i'}}\rangle  ={(\alpha_{2}+1)}/{[|\delta_{j'}^{(t)}|({|\delta_{j'}^{(t)}|}+\eta_{2})]}, } 
where $\delta_{j'}^{(t)}={\beta_{j'}}^{(t)}-{\beta_{j'+1}}^{(t)}$. In the M-step, the gradient w.r.t $\beta_{j}$ can be written as $D_{\beta_{j}}{Q} = \sum_{i=1}^{n}  \epsilon_{i,j} y_{i} -[\wt{\bd{\Lambda}}^{(t)}\bd{\beta}]_{j}$,   
where
\begin{equation*}
\resizebox{.99 \textwidth}{!} 
{
    $\wt{\bd{\Lambda}}^{(t)} =
\begin{bmatrix}
           \langle{\rho_{1}}\rangle+ \langle{\upsilon_{1}}\rangle, & -\langle{\upsilon_{1}}\rangle, &  &  &  & & & \\
            -\langle{\upsilon_{1}}\rangle, &\langle{\rho_{2}}\rangle+\langle{\upsilon_{1}}\rangle+\langle{\upsilon_{2}}\rangle, & -\langle{\upsilon_{2}}\rangle & & & && \\
                         & -\langle{\upsilon_{2}}\rangle, &\langle{\rho_{3}}\rangle+\langle{\upsilon_{2}}\rangle+\langle{\upsilon_{3}}\rangle, & -\langle{\upsilon_{3}}\rangle& && & \\
                                                                      & & &     & \ddots& &&\\
                                             & & &     & & -\langle{\upsilon_{m-2}}\rangle, &\langle{\rho_{m-1}}\rangle+\langle{\upsilon_{m-2}}\rangle+\langle{\upsilon_{m-1}}\rangle,& -\langle{\upsilon_{m-1}}\rangle \\
                                                                                          & & &  & &&-\langle{\upsilon_{m-1}}\rangle,& \langle{\rho_{m}}\rangle+\langle{\upsilon_{m-1}}\rangle\\
         \end{bmatrix}. 
 $
}
\end{equation*}

\paragraph{GDP-based PFL}
The GDP-based PFL penalty term is  
\eq{
\log{p(\bd{\beta})} & = \sum_{j=1}^{m}\log{p({\beta}_{j})} = \theta\sum_{j=1}^{m}\bigg[-\log{(2\xi_{1})}-(\alpha_{1}+1)\log{\bigg(1+\frac{|{\beta}_{j}|}{\alpha_{1} \xi_{1}}}\bigg)\bigg]\cr 
& + (1-\theta)\sum_{j<k}\bigg[-\log{(2\xi_{2})}-(\alpha_{2}+1)\log{\bigg(1+\frac{|{\beta}_{j}-{\beta}_{k}|}{\alpha_{2} \xi_{2}}}\bigg)\bigg],
}
where $\theta\in[0,1]$ is the weight parameter that balances selection with fusion. In the E-Step, we have
\eq{\langle{\rho_{j}}\rangle & ={\theta(\alpha_{1}+1)}/{[|{\beta_{j}}^{(t)}|({|{\beta_{j}}^{(t)}|}+\eta_{1})]}, \quad  \langle{\upsilon_{j,k}}\rangle  ={(1-\theta)(\alpha_{2}+1)}/{[|\delta_{j,k}^{(t)}|({|\delta_{j,k}^{(t)}|}+\eta_{2})]},
}
where $\delta_{j,k}^{(t)}={\beta_{j}}^{(t)}-{\beta_{k}}^{(t)}$. Similarly, in the M-step the gradient w.r.t $\beta_{j}$ can be written as $D_{\beta_{j}}{Q} = \sum_{i=1}^{n} \epsilon_{i,j} y_{i} -[\wt{\bd{\Lambda}}^{(t)}\bd{\beta}]_{j}$,  
where
\begin{equation*}
\resizebox{.99 \textwidth}{!} 
{
    $\wt{\bd{\Lambda}}^{(t)} =
\begin{bmatrix}
           \langle{\rho_{1}}\rangle+ \sum_{k\neq 1}\langle{\upsilon_{1,k}}\rangle, & -\langle{\upsilon_{1,2}}\rangle, & -\langle{\upsilon_{1,3}}\rangle, &  &  & \hdots & &-\langle{\upsilon_{1,m}}\rangle \\
            -\langle{\upsilon_{1,2}}\rangle, &\langle{\rho_{2}}\rangle+ \sum_{k\neq 2}\langle{\upsilon_{2,k}}\rangle, & -\langle{\upsilon_{2,3}}\rangle &  & & && -\langle{\upsilon_{2,m}}\rangle\\
                                                                  \vdots    & & &     & \ddots& &&\vdots \\
                                       -\langle{\upsilon_{1,m-1}}\rangle,       &  -\langle{\upsilon_{2,m-1}}\rangle,& &     & & -\langle{\upsilon_{m-2,m-1}}\rangle, &\langle{\rho_{m-1}}\rangle+ \sum_{k\neq (m-1)}\langle{\upsilon_{m-1,k}}\rangle,& -\langle{\upsilon_{m-1,m-1}}\rangle \\
                                            -\langle{\upsilon_{m-1,m}}\rangle,                                               &  & \hdots &  & &&-\langle{\upsilon_{m,m-1}}\rangle,& \langle{\rho_{m}}\rangle+ \sum_{k\neq m}\langle{\upsilon_{m,k}}\rangle\\
         \end{bmatrix}. 
 $
}
\end{equation*}

\vskip 0.2in
\bibliography{JMLR}

\end{document}